\documentclass[journal]{IEEEtran}

\usepackage{graphicx}
\usepackage{cite}
\usepackage{picinpar}
\usepackage{amsmath}
\usepackage{url}
\usepackage{flushend}
\usepackage[latin1]{inputenc}
\usepackage{colortbl}
\usepackage{soul}
\usepackage{multirow}
\usepackage{pifont}
\usepackage{color}
\usepackage{alltt}
\usepackage[hidelinks]{hyperref}
\usepackage{enumerate}
\usepackage{siunitx}
\usepackage{breakurl}
\usepackage{epstopdf}
\usepackage{pbox}

\usepackage{subfigure}
\usepackage{mathtools}
\usepackage{amssymb}

\usepackage{floatrow}
\newfloatcommand{capbtabbox}{table}[][\FBwidth]

\usepackage{bbding} 
\usepackage{pifont}
\newcommand{\cmark}{\ding{51}}

\DeclareMathOperator*{\minimize}{minimize}
\DeclareMathOperator*{\maximize}{maximize} 

\begin{document}
\title{	RaLL: End-to-end Radar Localization on Lidar Map Using Differentiable Measurement Model}

\author{
	\vskip 1em
	
	Huan Yin, \emph{Member, IEEE},
	Runjian Chen,
	Yue Wang, \emph{Member, IEEE},
	and Rong Xiong, \emph{Member, IEEE}
	
	\thanks{
		
		
		Huan Yin, Yue Wang, Runjian Chen and Rong Xiong are with the State Key Laboratory of Industrial Control and Technology, Zhejiang University, Hangzhou, China, and with the Institute of Cyber-Systems and Control, Zhejiang University, Hangzhou, China. (Corresponding author: Yue Wang, co-corresponding: Rong Xiong)
	}
}

\maketitle

\begin{abstract}
	Compared to the onboard camera and laser scanner, radar sensor provides lighting and weather invariant sensing, which is naturally suitable for long-term localization under adverse conditions. However, radar data is sparse and noisy, resulting in challenges for radar mapping. On the other hand, the most popular available map currently is built by lidar. In this paper, we propose an end-to-end deep learning framework for Radar Localization on Lidar Map (RaLL) to bridge the gap, which not only achieves the robust radar localization but also exploits the mature lidar mapping technique, thus reducing the cost of radar mapping. We first embed both sensor modals into a common feature space by a neural network. Then multiple offsets are added to the map modal for exhaustive similarity evaluation against the current radar modal, yielding the regression of the current pose. Finally, we apply this differentiable measurement model to a Kalman Filter (KF) to learn the whole sequential localization process in an end-to-end manner. \textit{The whole learning system is differentiable with the network based measurement model at the front-end and KF at the back-end.} To validate the feasibility and effectiveness, we employ multi-session multi-scene datasets collected from the real world, and the results demonstrate that our proposed system achieves superior performance over $90km$ driving, even in generalization scenarios where the model training is in UK, while testing in South Korea. We also release the source code publicly \footnote{The code of RaLL is available at \href{https://github.com/ZJUYH/RaLL}{https://github.com/ZJUYH/RaLL} and a video demonstration is available at \href{https://youtu.be/a3wEv-eVlcg}{https://youtu.be/a3wEv-eVlcg}.}.
\end{abstract}


\markboth{IEEE Transactions on Intelligent Transportation Systems}%
{}


\section{Introduction}
\IEEEPARstart{L}{ocalization} is a fundamental component for autonomous vehicles. A basic technique is to localize the vehicle via Global Positioning/Inertial Navigation System (GPS/INS) in outdoor scenes, which is infeasible in GPS-denied environments. Localization can also be accumulated by odometry data, while it is impracticable when the drift error becomes significant. Therefore, localization on pre-built map is indispensable, which tracks the robot pose with bounded uncertainty.

Recently many visual and lidar based approaches have been widely proposed in the field of robotics, but there still exist some difficulties in real applications. For instance, visual localization on visual map \cite{mur2015orb} and lidar map \cite{ding2019persistent} are still challenging when appearance changes significantly, like weather or season varies. For lidar localization, the technique is more mature, pose estimation can be achieved across day and night via point cloud registration \cite{krusi2015lighting, javanmardi2020pre}. Therefore, city scale lidar maps have been built for preparing commercial applications \cite{lu2019l3, barsan2018learning}. Unfortunately, in extreme weather conditions \cite{carballo2020libre}, like foggy and snowy, the lidar range measurements are very noisy, calling for the aid of other sensors.

To build a robust localization system, we consider radar is a good choice, which is naturally lighting and weather invariant. Therefore, radar odometry becomes a research focus recently \cite{2018ICRA_cen, 2019ICRA_cen,2019ICRA_aldera,Barnes2019MaskingByMoving,park2020pharao}, and radar Simultaneous Localization and Mapping (SLAM) \cite{hong2020radarslam}. A general idea is to build a map using radar, then localize the vehicle by aligning the radar measurements against the radar map. However, the raw radar data is noisy and sparse actually, and radar data processing is challenging for precise mapping, thus making  radar mapping and localization unapplicable. In addition, considering that the large-scale lidar map is available, we argue that repeating the whole mapping process for radar is extremely time and manual labor consuming, also bringing extra calibration between radar map and lidar map \cite{ma2019exploiting}.

\begin{figure}[t]
	\centering
	\includegraphics[width=9cm]{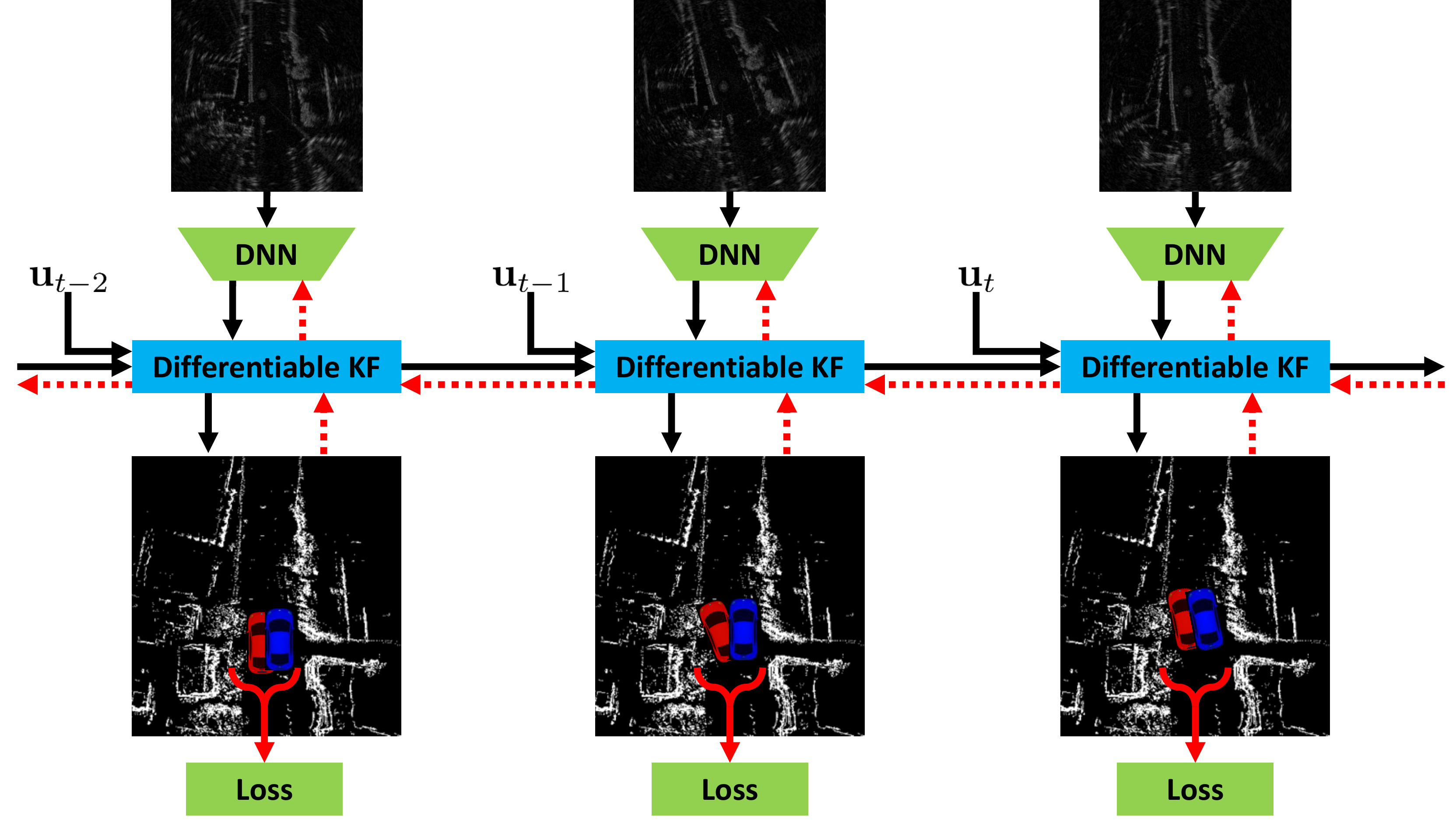}
	\caption{The schematics of RaLL. The blue and red vehicles stand for the estimated and ground truth poses respectively. The dashed line indicates the back-propagation of gradient from sequential loss to learnable deep network.}
	\label{teaser}	
\end{figure}

In this paper, we propose an end-to-end learning system to achieve radar based localization directly on lidar maps, as shown in Figure~\ref{teaser}. There are two challenges for RaLL. First, the common feature space of the two sensor modals is not explicitly supervised. We propose a network to learn the feature shared embeddings from the pose supervision by introducing a differentiable pose estimator to back-propagate the gradient. Second, after measuring the pose offset of radar and lidar, the uncertainty modeling of this measurement is not explicitly supervised by the network, preventing the probabilistic fusion between measurement and motion model. To address this problem, we apply the learned measurement model in a pose tracking Kalman Filter, enforcing the measurement uncertainty to be compatible to Gaussian fusion. By modeling the whole localization process as a sequential Gaussian distribution estimator, we can supervise the measurement uncertainty via maximizing likelihood. To evaluate the system performance, we utilize RobotCar dataset \cite{RadarRobotCarDatasetArXiv,maddern20171} for benchmark. In addition, the MulRan dataset \cite{kim2020mulran} is employed for testing the generalization of RaLL. In summary, the contributions of this paper are as follows:
\begin{itemize}
	\item A deep neural network architecture is proposed to learn the cross-modal shared feature embedding by back-propagating gradient from the pose supervision, which leverages the lidar map for radar localization.
	\item With the learned network being a differentiable measurement model, a Kalman Filter is proposed to generate fused estimation, which is learned in an end-to-end way to improve the accuracy.
	\item We conduct the model training and testing using RobotCar driving dataset (UK), and achieve similar experimental performance in Mulran dataset (South Korea), demonstrating good generalization of RaLL.
\end{itemize}

The rest of this paper is organized as follows: Section \ref{sec:related} reviews the related topics in recent years. The whole system is introduced in Section \ref{sec:methods}. Section \ref{sec:experiments} reports the experimental results on two datasets. We conclude a brief overview of our system and a future outlook in Section \ref{sec:conclusion}.



\begin{figure*}[t]
	\centering
	\includegraphics[width=\linewidth]{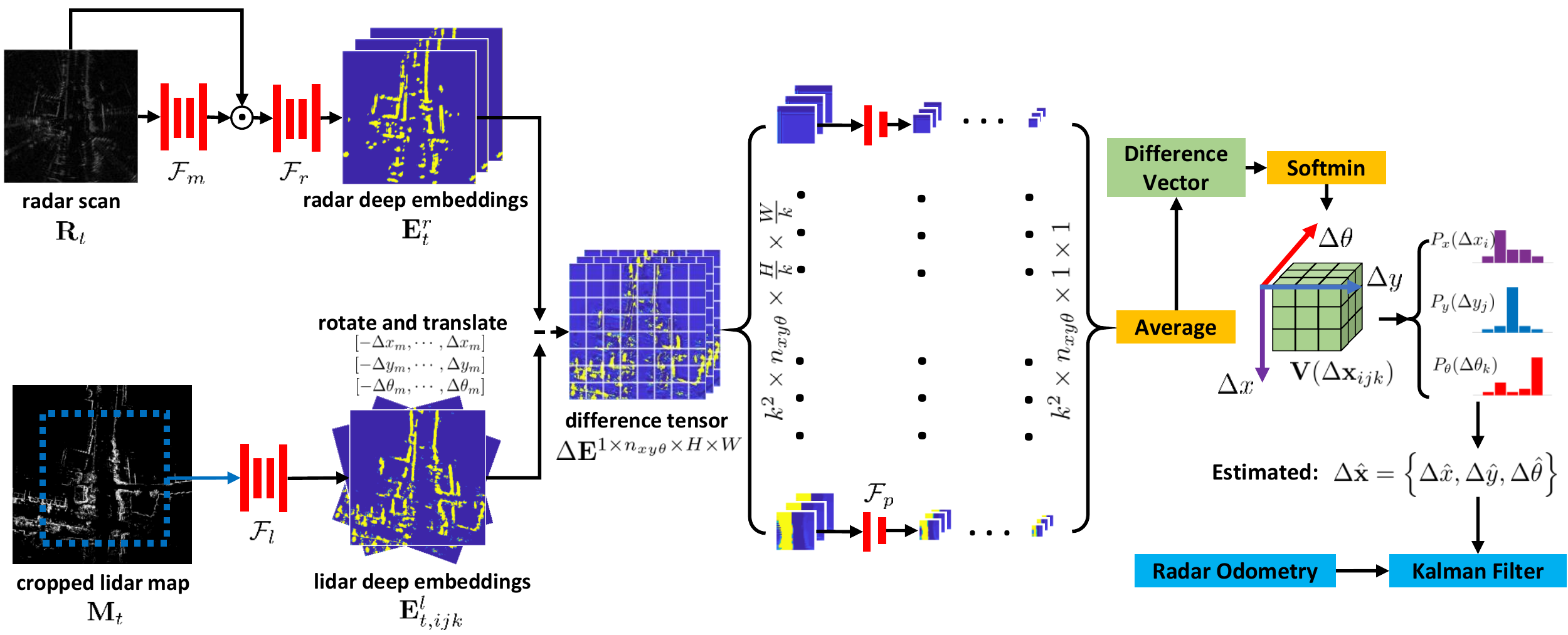}
	\caption{The architectural diagram of the proposed network RaLL. This neural network $\mathcal{F}$ consists of three U-Net $\mathcal{F}_m$, $\mathcal{F}_l$, $\mathcal{F}_r$ and one patch network $\mathcal{F}_p$. Finally, the probability distributions of offsets are obtained, which are the measurements of the following KF system for pose tracking.}
	\label{method}	
\end{figure*}

\section{Related Work}
\label{sec:related}

\subsection{SLAM and Localization on Pre-built Maps}
SLAM is widely used to construct maps in an unknown environment while simultaneously keeping track of the robot pose. Some visual SLAM \cite{ros2012visual, akilan2020multimodality} and lidar SLAM \cite{ding2018multi} were proposed to track the vehicle pose in the outdoor scenes. However, there remain some challenging problems in SLAM problem \cite{bresson2017simultaneous}, for example, the SLAM performance degradation with dynamics \cite{sun2017improving, sun2018motion} and the adverse weathers \cite{hong2020radarslam}.

Compared to the SLAM systems, localization on pre-built maps is a more desired technique, especially for long-term autonomy. The localization on pre-built maps avoids the online mapping session, thus making the localization more efficient. Some researchers proposed visual localization on visual maps \cite{huang2020gmmloc} and lidar maps \cite{ding2019persistent}, and also the lidar localization on lidar maps \cite{yin20193d}. However, visual or lidar localization is still challenging in adverse weathers. Compared to these two sensors, radar sensor provides lighting and weather invariant sensing. So in this paper, we propose a radar localization on pre-built lidar maps, in which high definition lidar maps are constructed offline and then radar localization is performed.

\subsection{Localization Using Radar Sensor}
Radar sensor has been applied for various perception tasks widely in autonomous driving area, mainly focused on dynamic objects detection on roads. Some researchers achieved indoor positioning using low-cost radar sensors, including localization on Computer Aided Design (CAD) model \cite{park2019radar} or laser map \cite{lu2020see}. But in challenging outdoor scenarios, precise radar localization is blocked by diverse types of sensor noises and Doppler velocity, therefore, most of studies are based on data pre-processing and sensor modeling \cite{vivet2013localization, schuster2016robust, schuster2016landmark, rapp2017probabilistic}, or combined with other sensors on vehicles \cite{narula2020automotive}.

Recently the Navtech Frequency Modulated Continuous Wave
(FMCW) radar sensor brings less Doppler effect, higher resolution and 360$^\circ$-view in data collection \cite{RadarRobotCarDatasetArXiv,maddern20171,kim2020mulran, hong2020radarslam, burnett2020we}. The development of this sensor technique results in new approaches to robotic localization research. For place recognition, part of positioning system, researchers extended the existing methods in vision or lidar community to radar data \cite{gadd2020lookaroundyou, suaftescu2020kidnapped, kim2020mulran}. As for ego-motion estimation, sparse landmarks and keypoints are extracted manually to build efficient data association model \cite{2018ICRA_cen, 2019ICRA_cen}, and these salient keypoints or pixels can also be learned by neural networks in \cite{2019ICRA_aldera, UnderTheRadarICRA2020}. The state-of-the-art odometric performance was presented in \cite{Barnes2019MaskingByMoving}, in which an end-to-end radar odometry system was proposed with the supervision of ground truth poses.

Considering the unavoidable drift in odometry, the registration between live scan and prior map is still essential for long-term localization. Hong et al. \cite{hong2020radarslam} used conventional SLAM algorithms to build a radar mapping system. Localization with multiple sensor modalities was presented in \cite{tang2020rsl}, in which radar data is localized against satellite images with a coarse initial estimate. The authors then presented more accurate results using multiple networks in \cite{tang2020self}. In \cite{yin2020radar}, to connect radar scan and lidar map, a Generative Adversarial Network (GAN) was used to transfer the radar to fake lidar data, and then Monte Carlo Localization (MCL) was applied for pose tracking.

\subsection{Deep Learning based Localization}
With the advances of machine learning, some geometric or theoretical problems of robot localization and mapping were solved by data-driven \cite{yin20203d, huang2020monocular, chen2020survey}. A direct way is to localize the sensor by end-to-end learning without geometric information \cite{kendall2015posenet}, which requires large training data to avoid overfitting. Some researchers \cite{detone2018superpoint} established feature correspondences by deep learning and used traditional solvers for pose estimation, while these solvers are indifferentiable for end-to-end learning.

This paper is inspired by several recent deep learning based approaches \cite{barsan2018learning, Barnes2019MaskingByMoving, lu2019l3}. In these research works, localization is performed on same sensor modalities. While in this paper, we model the similarity measure and pose regression on the different modalities via deep neural networks. We also train and test the networks with the sequential data, thus building an end-to-end localization system in time domain.


\section{Methods}
\label{sec:methods}

The framework of RaLL is a differentiable Kalman Filter embedded with a neural network based measurement model, which is introduced in Section \ref{sec:KF}. The network architecture of the measurement model consists of two stages. The first stage embeds the radar scan and lidar map into a common feature space, which is introduced in Section \ref{sec:embeddings}. The second stage evaluates the similarity of the features extracted from two branches above, and yields the final estimation, which is introduced in Section \ref{sec:similarity}. Finally, we show the training strategy for RaLL in Section \ref{sec:impl}.


\subsection{Comparable Feature Embeddings}
\label{sec:embeddings}

We represent both radar scan and lidar map as bird-eye view images. At time $t$, one scan $\mathbf{R}_t^{H\times W}$ is generated from the radar, and an initial estimation of the pose is $\overline{\mathbf{x}}_t \in SE(2)$. For the prior 2D laser map $\mathbf{M}$, we crop it to $\mathbf{M}_t^{H\times W}$ at pose  $\overline{\mathbf{x}}_t$. With the input $\mathbf{R}_t$ and $\mathbf{M}_t$, a neural network $\mathcal{F}$ is designed to estimate the offset $\Delta \mathbf{x}=\left\lbrace \Delta x, \Delta y, \Delta \theta\right\rbrace $ between $\overline{\mathbf{x}}_t$ and the ground truth pose $\mathbf{x}_t$. Therefore, the ground truth offset $\Delta \mathbf{x}$ is the supervision for training $\mathcal{F}$.

There are three differences between the data from radar and lidar map, which prevents the direct alignment. First, various types of noises exist in radar data, while the lidar map is much cleaner. Second, there are occlusions in the current scan, while the lidar map is more completed because of the multi-frame integration. Finally, the raw data forms are different: $\mathbf{R}_t \in [0,1]$ has continuous radar intensities while $\mathbf{M}_t$ are represented by binary, or lidar grids. We embed the two heterogeneous modalities into a comparable feature space by training a deep neural network.

\begin{figure*}[t]
	\centering
	\includegraphics[width=\linewidth]{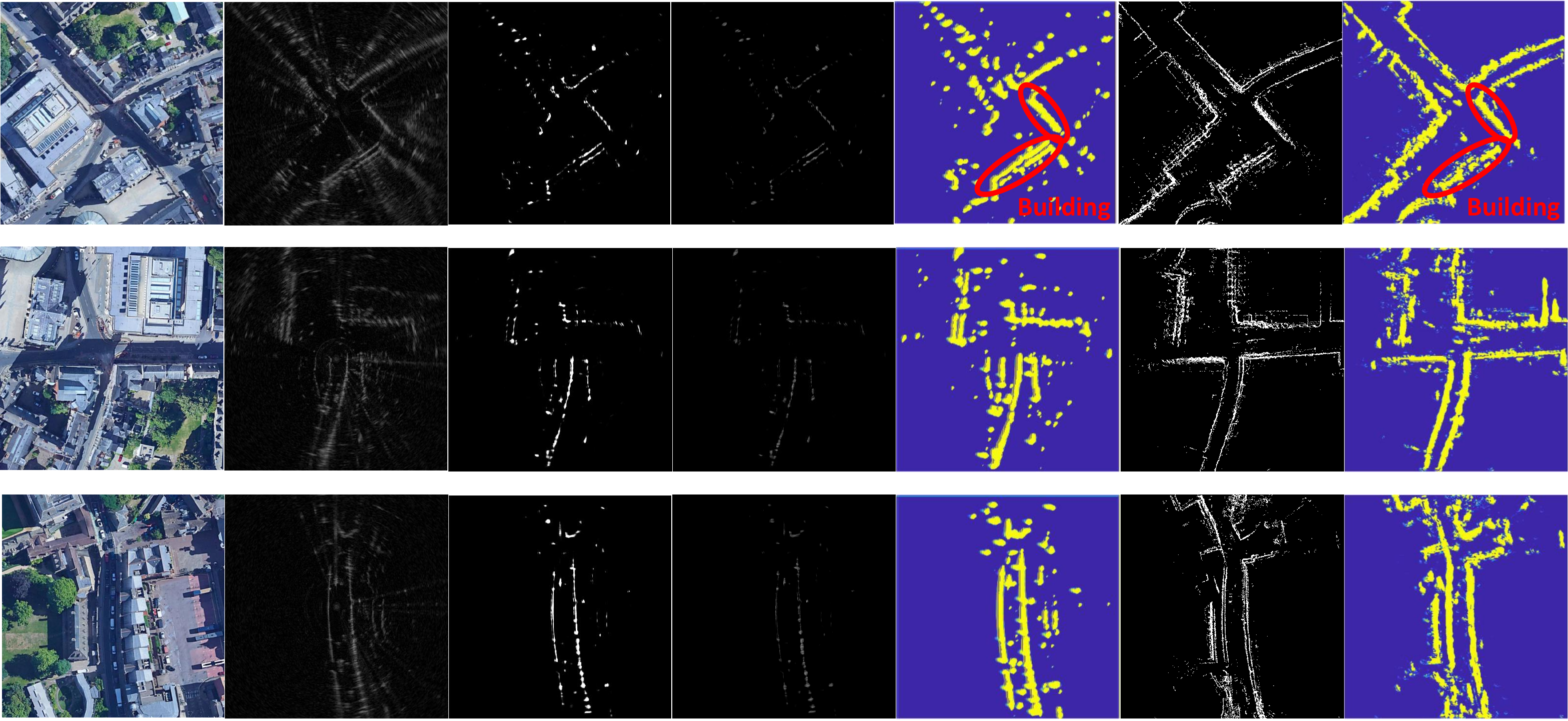}
	\caption{The feature embeddings from networks. From top to bottom: three scenes in RobotCar dataset. From left to right: satellite image, $\mathbf{R}_t$, $\mathcal{F}_m(\mathbf{R}_t)$, $\mathbf{R}_t^m$, $\mathbf{E}_t^r$; then followed $\mathbf{M}_t$ and $\mathbf{E}_t^l$. For better understanding, we annotate the buildings on the feature embeddings of of first row.}
	\label{features}	
\end{figure*}

We establish a two-stream neural network for radar and lidar, shown in Figure~\ref{method}. The whole network consists of three U-Net architectures \cite{ronneberger2015u}, denoted as $\mathcal{F}_m$, $\mathcal{F}_r$ and $\mathcal{F}_l$. To suppress the noise in radar, we apply a masking network $\mathcal{F}_m$ to learn a noise mask as in \cite{Barnes2019MaskingByMoving}. The sigmoid activation in the final layer constrains the masking output range to $[0,1]$, then a de-noised representation $\mathbf{R}_t^m$ is obtained by element-wise multiplication $\mathbf{R}_t^m = \mathbf{R}_t \odot \mathcal{F}_m(\mathbf{R}_t)$. We notice that the filtered $\mathbf{R}_t^m$ is much clearer as shown in Figure~\ref{features}. Then we learn the feature embedding which is comparable between the de-noised radar scan and the lidar map: $\mathcal{F}_r(\mathbf{R}_t^m) \mapsto \mathbf{E}_t^r$, $\mathcal{F}_l(\mathbf{M}_t) \mapsto \mathbf{E}_t^l$. We set the final activation function as ReLU in $\mathcal{F}_r$ and $\mathcal{F}_l$ for feature extraction. Figure~\ref{features} demonstrates several learned deep embeddings. It is clear that the features salient in both modalities, are directly comparable in the feature embedding space, such as buildings facade etc.

\subsection{Similarity-based Offset Estimation}
\label{sec:similarity}

Given the initial pose $\overline{\mathbf{x}}_t$, the pose offset should lie in a limited interval, denoted as $- \Delta x_m \leqslant \Delta x \leqslant \Delta x_m, \Delta x_m > 0$; the same for $\Delta y$ and $\Delta \theta$ in the range $[-\Delta y_m,\Delta y_m]$ and $[-\Delta \theta_m,\Delta \theta_m]$. Then we divide the 3D solution space for the offset into grids with certain resolutions $\delta x,\delta y,\delta \theta$, and there are $n_{xy\theta} = n_x \times n_y \times n_{\theta}$ candidates in total, e.g. $n_x=\frac{\Delta x_m - ( - \Delta x_m)}{\delta x} + 1$. Hence each candidate in the solution space is denoted as $\Delta \mathbf{x}_{ijk}= \left\lbrace \Delta x_i, \Delta y_j, \Delta \theta_k \right\rbrace $, where $1 \leqslant i \leqslant n_x, 1 \leqslant j \leqslant n_y, 1 \leqslant k \leqslant n_\theta$. As a result, the absolute pose of each candidate is calculated by transforming the initial pose $\mathbf{x}_t$ using the offset $\Delta \mathbf{x}_{ijk}$ as follow:
\begin{equation}\label{disturbance}
\overline{\mathbf{x}}_t \boxplus \Delta \mathbf{x}_{ijk} =
\begin{bmatrix} \overline{x}_t + \cos(\overline{\theta}_t + \Delta \theta_k)\Delta x_i - \sin(\overline{\theta}_t + \Delta \theta_k)\Delta y_j \\
\overline{y}_t + \sin(\overline{\theta}_t + \Delta \theta_k)\Delta x_i + \cos(\overline{\theta}_t + \Delta \theta_k)\Delta y_j \\
\overline{\theta}_t + \Delta \theta_k \end{bmatrix},
\end{equation}
where $\boxplus$, $\overline{\mathbf{x}}_t$ and $\Delta \mathbf{x}_{ijk}$ stand for transformation process, the current pose and pose offset respectively. 

At the level of feature embeddings, if the vehicle rotates and translates with the offset $\Delta \mathbf{x}_{ijk}$, the surrounding map will change respectively. In this context, we rotate and translate the map embedding $\mathbf{E}_t^l$ using $n_{xy\theta}$ candidates, resulting in $n_{xy\theta}$ map blocks $\mathbf{E}^l_{t,ijk}$. Specifically, for every element $\left[ e_u, e_v \right] $ in $\mathbf{E}_t^l$, the transformed element $\left[ e_{u,ijk}, e_{v,ijk} \right] $ in $\mathbf{E}^l_{t,ijk}$ is calculated with the offset $\Delta \mathbf{x}_{ijk}$ as follows:
\begin{equation}
\begin{bmatrix}
e_{u,ijk}\\e_{v,ijk}\\1
\end{bmatrix} = 
\begin{bmatrix}
cos(\Delta \theta_k) & sin(\Delta \theta_k) & -\Delta y_j\\
-sin(\Delta \theta_k) & cos(\Delta \theta_k) & -\Delta x_i\\
0 & 0 &1
\end{bmatrix}
\begin{bmatrix}
e_{u}\\e_{v}\\1
\end{bmatrix},
\end{equation}
where $\Delta x_i$, $\Delta y_j$ and $\Delta \theta_k$ are variables in offset $\Delta \mathbf{x}_{ijk}$.

\begin{figure}[t]
	\centering
	\includegraphics[width=8cm]{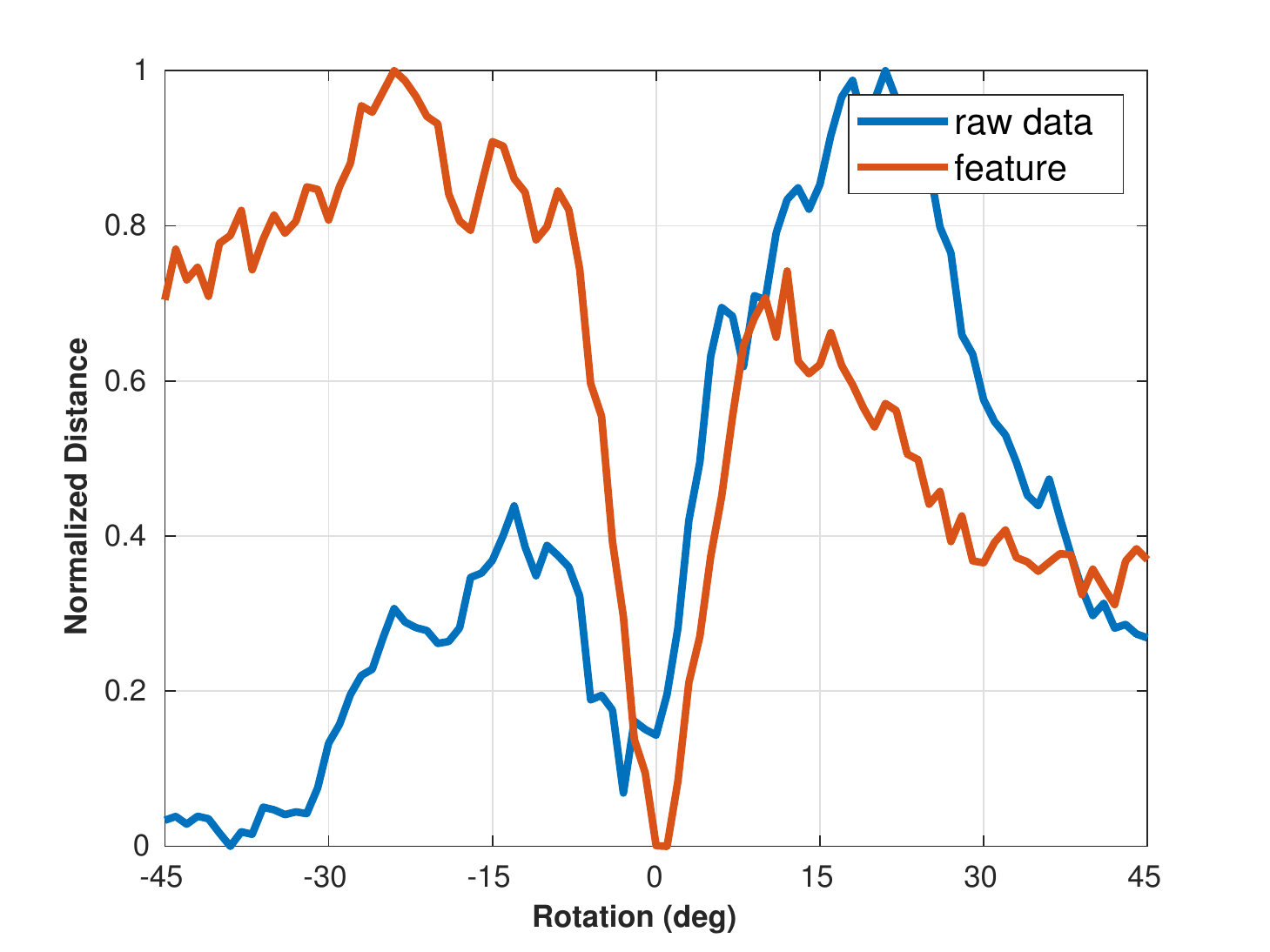}
	\caption{The comparable analysis of raw data and feature embeddings. We rotate the lidar map and the learned embedding around the ground truth, and then measure the normalized distance compared to the radar respectively.}
	\label{rotation}	
\end{figure}

As shown in Figure~\ref{features}, after the feature embeddings $\mathbf{E}_t^r$ and $\mathbf{E}_{t,ijk}^l$ are learned, we suppose the features should be more comparable than the raw data by the element-wise subtraction $(\mathbf{E}_t^r - \mathbf{E}_{t,ijk}^l)$. To validate this hypothesis, we rotate the raw map and embeddings of the first scene in Figure~\ref{features}, and then measure the distance compared to the radar data and embedding respectively. The normalized distance in Figure~\ref{rotation} shows that the substraction of feature embeddings achieves the minimum cost at the ground truth pose, indicating that the learned features are more comparable. Repeating the subtraction for each candidate, we form a difference tensor as $\Delta \mathbf{E}$ with the size $n_{xy\theta} \times H\times W$, which is the start point for the following similarity learning.

Considering the occlusion, we further divide $\Delta \mathbf{E}$ into $k \times k$ small patches as shown in Figure~\ref{method}. For each patch, we utilize a patch network $\mathcal{F}_p$ to derive a patch based difference. Finally, we average the $k^2$ patch based differences and obtain a difference score between the radar scan feature $\mathbf{E}^r_{t}$ and each candidate $\mathbf{E}^l_{t,ijk}$, leading to a difference vector with a length of $n_{xy\theta}$.

Essentially, if a candidate pose, say $(\overline{\mathbf{x}}_t \boxplus \Delta \mathbf{x}_{ijk} )$ is close to the ground truth pose $\mathbf{x}_t$, $\mathbf{E}_t^r$ should be similar with $\mathbf{E}_{t,ijk}^l$, hence leading to a low difference at the $ijk$th position of the difference vector. As the lower the difference is, the closer the candidate and the ground truth lie, we take a softmin module to transform and normalize the difference vector to form a probability distribution. Then, we reshape the distribution into a 3-dimensional cost volume $\mathbf{V}(\Delta \mathbf{x}_{ijk})$. In summary, from the current radar scan and the lidar map, the proposed network produces a probability distribution of the pose offset to be estimated, which states as follows:
\begin{equation}\label{network}
\mathcal{F}_{m,r,l,p}(\mathbf{R}_t, \mathbf{M}, \overline{\mathbf{x}}_t, \Delta \mathbf{x}_{ijk}) \mapsto \mathbf{V}(\Delta \mathbf{x}_{ijk}),
\end{equation}
where $\mathcal{F}_{m,r,l,p}$ is the neural network and $\mathbf{V}$ is the cost volume.

We then sum over the values along each axis to derive the marginal distribution:


\begin{equation}\label{Prob_xyt}
\left\{  
\begin{array}{lr}  
P_x(\Delta x_i)= \sum\limits_{\Delta y_j,\Delta \theta_k} \mathbf{V}(\Delta \mathbf{x}_{ijk}) \\  
P_y(\Delta y_j)= \sum\limits_{\Delta x_i,\Delta \theta_k} \mathbf{V}(\Delta \mathbf{x}_{ijk}) \\  
P_\theta(\Delta \theta_k)= \sum\limits_{\Delta x_i,\Delta y_j} \mathbf{V}(\Delta \mathbf{x}_{ijk})
\end{array},  
\right.  
\end{equation}
in which the probabilities $P_x(\Delta x_i)$, $P_y(\Delta y_j)$ and $P_\theta(\Delta \theta_k)$ are obtained along the axes in the cost volume $\mathbf{V}$.

Upon the marginal distribution, we design two losses to train the network: $\mathcal{L}_1 + \mathcal{L}_2$. When the estimation problem is regarded as three separated classification problems of $\left\lbrace \Delta x_i, \Delta y_j, \Delta \theta_k \right\rbrace$, a cross entropy loss can be formed as follows:
\begin{equation}\label{L_1}
\mathcal{L}_1 =-\sum_{\Delta x_i}C_x\log(P_x)
- \sum_{\Delta y_i} C_y\log(P_y)
-\sum_{\Delta \theta_k} C_\theta\log(P_\theta),
\end{equation}
where $C_x$, $C_y$ and $C_\theta$ are one hot encodings of the ground truth, which is the supervision. On the other hand, as a regression problem, we utilize the marginal expectations of the three marginal distributions as the estimators:
\begin{equation}\label{estimated}
\Delta \hat{x} = \sum_{\Delta x_i}P_x\cdot\Delta x_i,
\Delta \hat{y} = \sum_{\Delta y_j}P_y\cdot\Delta y_j,
\Delta \hat{\theta} = \sum_{\Delta \theta_k}P_\theta\cdot\Delta \theta_k,
\end{equation}
where the estimated pose offset $\Delta \hat{\mathbf{x}} = \left\lbrace  \Delta \hat{x}, \Delta \hat{y}, \Delta \hat{\theta} \right\rbrace $ is obtained. Then the second loss is constructed by the squared error between $\Delta \hat{\mathbf{x}}$ and ground truth $\Delta\mathbf{x} = \left\lbrace  \Delta x, \Delta y, \Delta \theta \right\rbrace$ as follows:
\begin{equation}\label{L_2}
\mathcal{L}_2 = (\Delta \hat{x} - \Delta x)^2 + (\Delta \hat{y} - \Delta y)^2 + \alpha \cdot (\Delta \hat{\theta} - \Delta \theta)^2,
\end{equation}
where $\alpha$ is a constant value to balance the translation metric $(m)$ and rotation angle metric $(^\circ)$.

\subsection{Differentiable Kalman Filter}
\label{sec:KF}

Kalman Filter is a general technique to fuse the sensor data in sequential. It consists of two steps, the prediction step using motion model and then followed by the updating step using measurement model, formulating a Bayesian recursive estimator with both models having Gaussian distributions. 

Firstly, we propose an Iterative Closest Point (ICP) based radar odometry as the motion model, which estimates the relative pose between two timestamps i.e. $\mathbf{u}_{t-1}$ from $\mathbf{R}_{t-1}$ to $\mathbf{R}_t$. Specifically, radar points with high intensities are extracted from the raw data, and an intensity threshold is used to select salient points. Then ICP \cite{besl1992method} is performed on these filtered points to estimate the pose. Based on this radar odometry, the prediction step of is formulated as follows:
\begin{equation} \label{prediction}
\begin{split}
&\overline{\mathbf{x}}_t = f(\hat{\mathbf{x}}_{t-1}, \mathbf{u}_{t-1}) \\
&\overline{\mathbf{\Sigma}}_t = \mathbf{F}_t \hat{\mathbf{\Sigma}}_{t-1} \mathbf{F}_t^{\rm T} + \mathbf{\Sigma}_m
\end{split},
\end{equation}
where $f(\cdot)$ accumulates the ego-motion on the previous estimated pose, and $\mathbf{F}_t$ is the Jacobian of $f(\cdot)$. The predicted pose $\overline{\mathbf{x}}_t$ is calculated with the motion model in this equation. The covariance $\overline{\mathbf{\Sigma}}_t$ is also propagated and inserted with the odometry covariance $\mathbf{\Sigma}_m$, which is a pre-defined coefficient.

\begin{figure}[t]
	\centering
	\includegraphics[width=9cm]{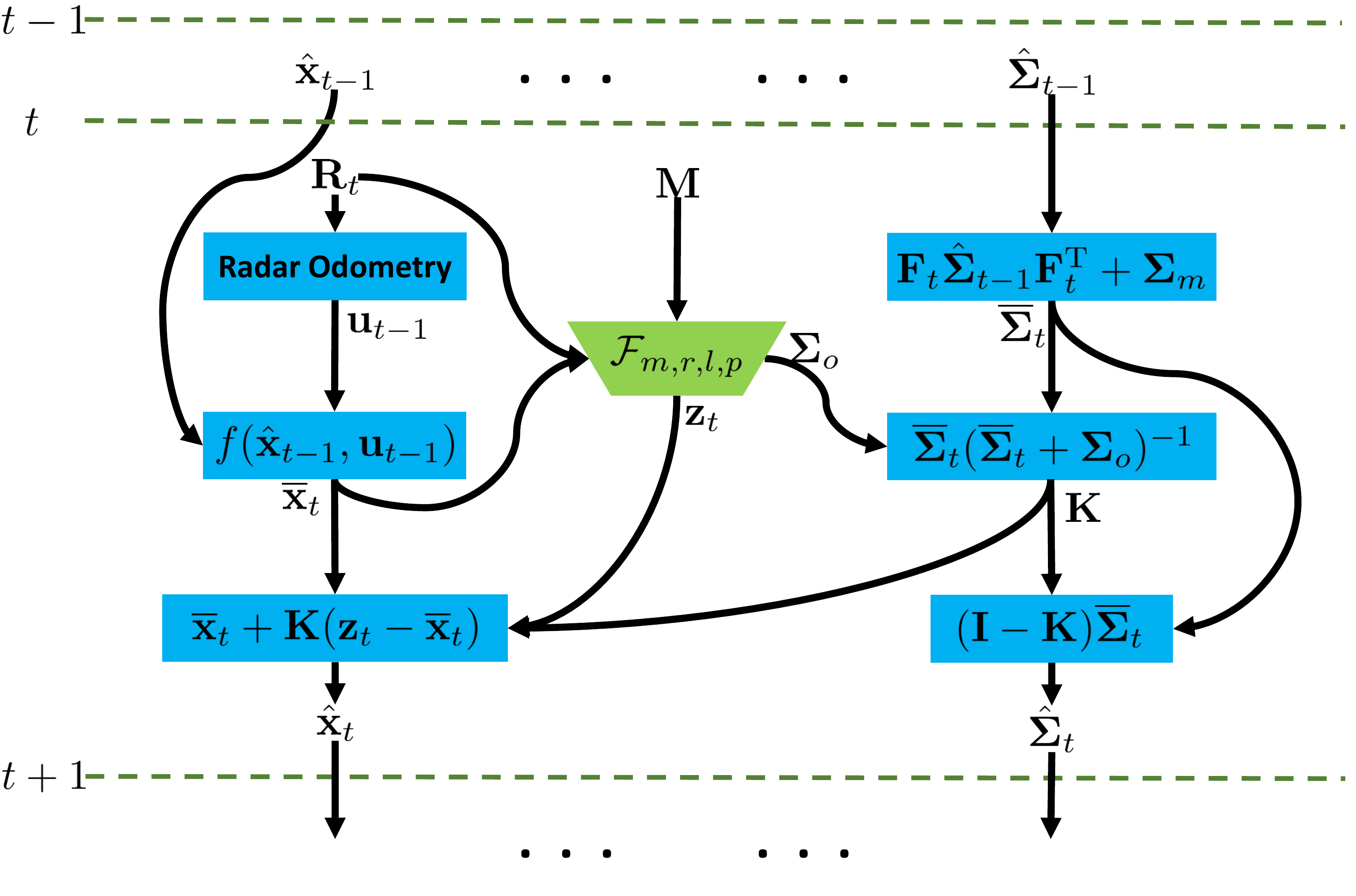}
	\caption{The computation graph of KF. The neural network is regarded as the measurement model.}
	\label{EKF}	
\end{figure}

In the update step, we utilize the networks $\mathcal{F}_{m,r,l,p}$ as the measurement model. We estimate the offset $\Delta \mathbf{x}_t$ at the predicted location $\overline{\mathbf{x}}_t$, then a global observation is generated by applying $\mathbf{z}_t= \overline{\mathbf{x}}_t \boxplus \Delta \mathbf{x}_t$, where $\boxplus$ is the pose transformation process in Equation \ref{disturbance}. With this GPS-like observation $\mathbf{z}_t$, we can derive the measurement model. We also calculate the observation covariance $\mathbf{\Sigma}_o$ according to the probability distributions $P_x, P_y$ and $P_\theta$ derived above. Overall, the update step is formulated as follows:
\begin{equation} \label{prediction}
\begin{split}
&\mathbf{K} = \overline{\mathbf{\Sigma}}_t  ( \overline{\mathbf{\Sigma}}_t  + \mathbf{\Sigma}_o)^{-1} \\
&\hat{\mathbf{x}}_t = \overline{\mathbf{x}}_t + \mathbf{K}(\mathbf{z}_t -\overline{\mathbf{x}}_t)\\
&\hat{\mathbf{\Sigma}}_t = (\mathbf{I} - \mathbf{K})\overline{\mathbf{\Sigma}}_t
\end{split},
\end{equation}
where the $\mathbf{K}$, $\hat{\mathbf{x}}_t$ and $\hat{\mathbf{\Sigma}}_t$ stand for the Kalman gain, the estimated pose and the covariance respectively. The network aided KF system is presented in Figure~\ref{EKF}. As both models are differentiable, the filter can be regarded as a recurrent network, which means that the measurement model can be trained by back-propagating gradients from future steps. 

At the probabilistic perspective, KF generates the Gaussian posterior of the 3-dimensional pose in sequential, which is  $\mathcal{N}(\hat{\mathbf{x}}_t, \hat{\mathbf{\Sigma}}_t)$. Therefore, with the sequential ground truth pose $\{\mathbf{x}_t\}$ available, we apply the maximum likelihood to tune the network parameters as follows:
\begin{equation}
\begin{split}
&~~~ \maximize~ \mathcal{N}(\mathbf{x}_t;~\hat{\mathbf{x}}_t, \hat{\mathbf{\Sigma}}_t) \\
&= \max~\frac{1}{((2\pi)^3{\rm det}(\hat{\mathbf{\Sigma}}_t))^{\frac{1}{2}}}{\rm exp}(-\frac{1}{2}(\hat{\mathbf{x}}_t - \mathbf{x}_t)^{\rm T}\hat{\mathbf{\Sigma}}_t^{-1}(\hat{\mathbf{x}}_t - \mathbf{x}_t)),
\end{split}
\end{equation}
where $\rm det(\cdot)$ is the determinant of a matrix and we apply negative log-likelihood to formulate the minimization as
\begin{equation}
\minimize~\frac{1}{2}{\rm log}((2\pi)^3{\rm det}(\hat{\mathbf{\Sigma}}_t)) + \frac{1}{2}(\hat{\mathbf{x}}_t - \mathbf{x}_t)^{\rm T}\hat{\mathbf{\Sigma}}_t^{-1}(\hat{\mathbf{x}}_t - \mathbf{x}_t).
\end{equation}

Based on the analysis, by discarding the constant terms, we can simply derive the third loss $\mathcal{L}_3$ for end-to-end training of the sequential localization as
\begin{equation}\label{L_3}
\mathcal{L}_3 = \frac{1}{k}\sum_{t-k}^{t} (\hat{\mathbf{x}}_t - \mathbf{x}_t)^{\rm T}\hat{\mathbf{\Sigma}}_t^{-1}(\hat{\mathbf{x}}_t - \mathbf{x}_t) + \beta \cdot {\rm det}(\hat{\mathbf{\Sigma}}_t),
\end{equation}
where $\beta$ is a constant balance factor, and $k$ denotes the length of sequence for training. This loss forms a balance between the likelihood and a lower uncertainty in the posterior.

\subsection{Implementation and Training Strategy}
\label{sec:impl}

We implement the proposed RaLL using PyTorch \cite{paszke2019pytorch}, including networks at the front-end and KF at the back-end. The network architecture is presented in Appendix of this paper. As for the Differentiable Kalman Filter (DKF) proposed in Section~\ref{sec:KF}, it can be regarded as a block with feed-forward and back-propagation process. The whole system is trained on a single Nvidia Titan X GPU. As for the motion model, the ICP based radar odometry is implemented by using libpointmatcher \cite{pomerleau2013comparing}.

We first train the measurement model $\mathcal{F}_{\mathcal{L}_1 + \mathcal{L}_2}$ using the single step loss $\mathcal{L}_1 + \mathcal{L}_2$. The training and test data is augmented by randomly sampling the initial poses near the ground truth poses. Furthermore, the pre-trained network is then trained with the sequential loss $\mathcal{L}_3$ and differentiable Kalman Filter in an end-to-end manner, denoted as $\mathcal{F}_{\mathcal{L}_1 + \mathcal{L}_2, \mathcal{L}_3}$. The input data for training $\mathcal{F}_{\mathcal{L}_1 + \mathcal{L}_2, \mathcal{L}_3}$ is sequences of temporal radar scans and corresponding ground truth poses. The performance tests with the trained $\mathcal{F}_{\mathcal{L}_1 + \mathcal{L}_2}$ and  $\mathcal{F}_{\mathcal{L}_1 + \mathcal{L}_2, \mathcal{L}_3}$ are presented in Section~\ref{sec:ablation} and \ref{sec:tracking}.


\section{Experiments}
\label{sec:experiments}

In Section \ref{sec:dataset}, we first introduce the datasets for evaluation, then followed an ablation study in Section \ref{sec:ablation}. The performance of pose tracking result is presented in Section \ref{sec:tracking}, with comparisons to competitive methods. We also present the performance in dynamic environments in Section \ref{sec:dynamic} and the system operating efficiency in Section \ref{sec:efficiency}. Finally, the proposed measurement model is compared to recent localization methods with large inital offsets.

\begin{table}[t]
	\renewcommand\arraystretch{1.2}
	\begin{center}
		\caption{Sequences 01-06 of RobotCar Dataset at Oxford}
		\label{RobotCar_Dataset}
		\begin{tabular}{p{1.6cm}<{\centering}|p{1.4cm}<{\centering}|p{4.6cm}<{\centering}}
			\hline
			\hline
			Date& Len.(km)& Sequence \\
			\hline
			10/01/2019& 9.02& Oxford-01: mapping \& training \\
			10/01/2019& 9.04& Oxford-02: localization test \\
			11/01/2019& 9.03& Oxford-03: localization test \\
			16/01/2019& 9.01& Oxford-04: localization test \\
			17/01/2019& 9.02& Oxford-05: localization test \\
			18/01/2019& 9.01& Oxford-06: localization test \\
			\hline
			\hline
		\end{tabular}
	\end{center}
\end{table}

\begin{table}[!t]
	\renewcommand\arraystretch{1.2}
	\begin{center}
		\caption{Sequences of MulRan Dataset at KAIST, DCC and Riverside}
		\label{MulRan_Dataset}
		\begin{tabular}{p{1.6cm}<{\centering}|p{1.4cm}<{\centering}|p{4.5cm}<{\centering}}
			\hline
			\hline
			Date& Len.(km)& Sequence (Generalization) \\
			\hline
			02/08/2019& 4.91& DCC-01: mapping \& test \\
			23/08/2019& 4.27& DCC-02: localization test \\
			03/09/2019& 5.42& DCC-03: localization test \\
			\hline
			20/06/2019& 6.13& KAIST-01: localization test \\
			23/08/2019& 5.97& KAIST-02: mapping \& test \\
			02/09/2019& 6.25& KAIST-03: localization test \\
			\hline
			02/08/2019& 7.25& Riverside-01: localization test \\
			16/08/2019& 6.61& Riverside-02: mapping \& test \\
			\hline
			\hline
		\end{tabular}
	\end{center}
\end{table}

\subsection{Datasets}
\label{sec:dataset}

We conduct the experiments on the public autonomous driving datasets, Oxford Radar RobotCar (RobotCar)\footnote{https://oxford-robotics-institute.github.io/radar-robotcar-dataset} \cite{RadarRobotCarDatasetArXiv,maddern20171} in UK and Multimodal Range Dataset (MulRan)\footnote{https://sites.google.com/view/mulran-pr}  \cite{kim2020mulran} in South Korea. Both of these datasets include ground truth poses, extrinsic calibration, raw lidar and radar data. The sensors types and locations on vehicles are different in the two datasets, thus validating the generalization of the proposed method indirectly.

\begin{figure*}[t]
	\centering
	\subfigure[Train and test sections]{	\label{train-test}
		\includegraphics[width=5.6cm]{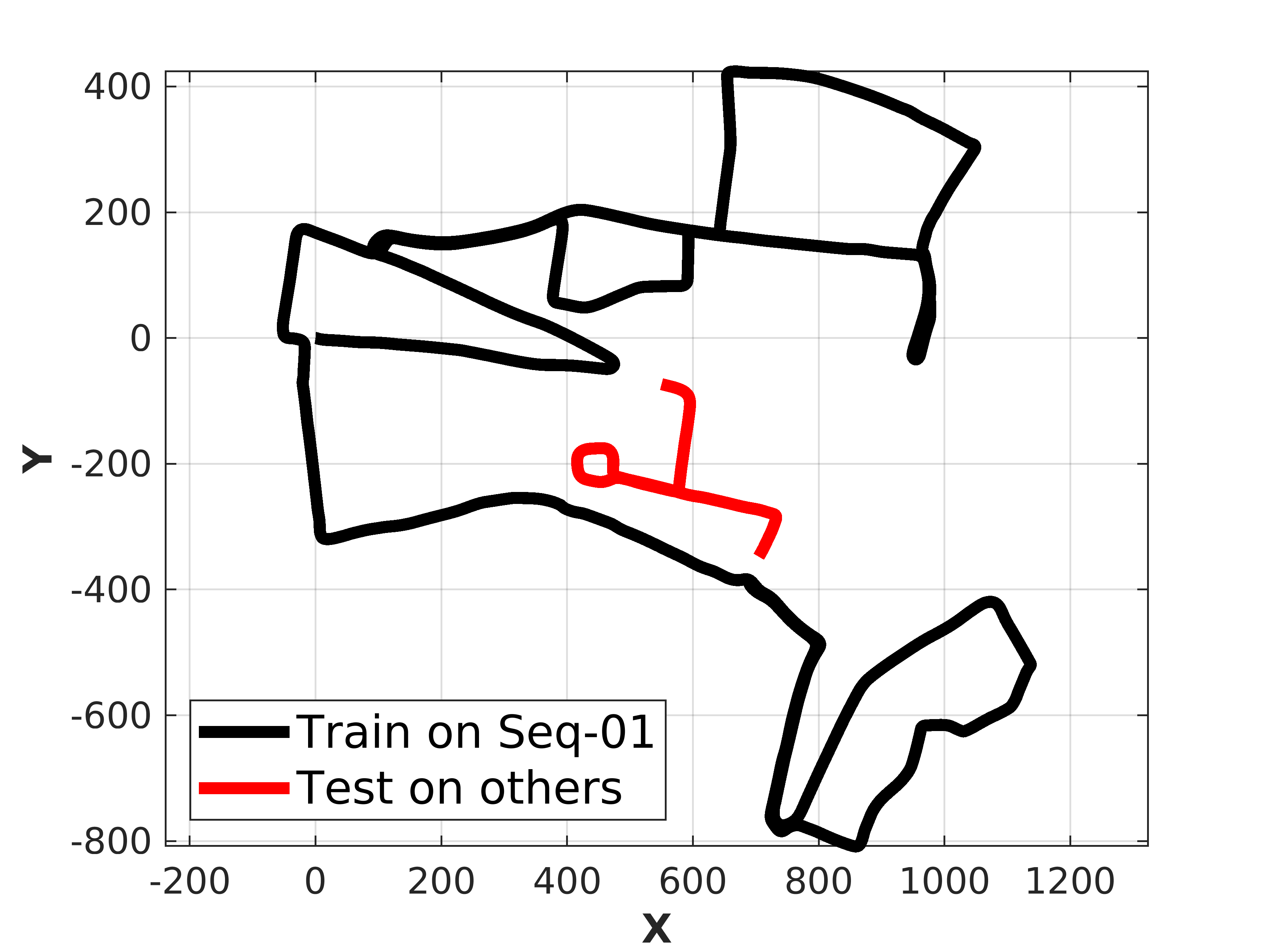}}
	\subfigure[Localization trajectories on Oxford-02]{	\label{Seq-02}
		\includegraphics[width=5.8cm]{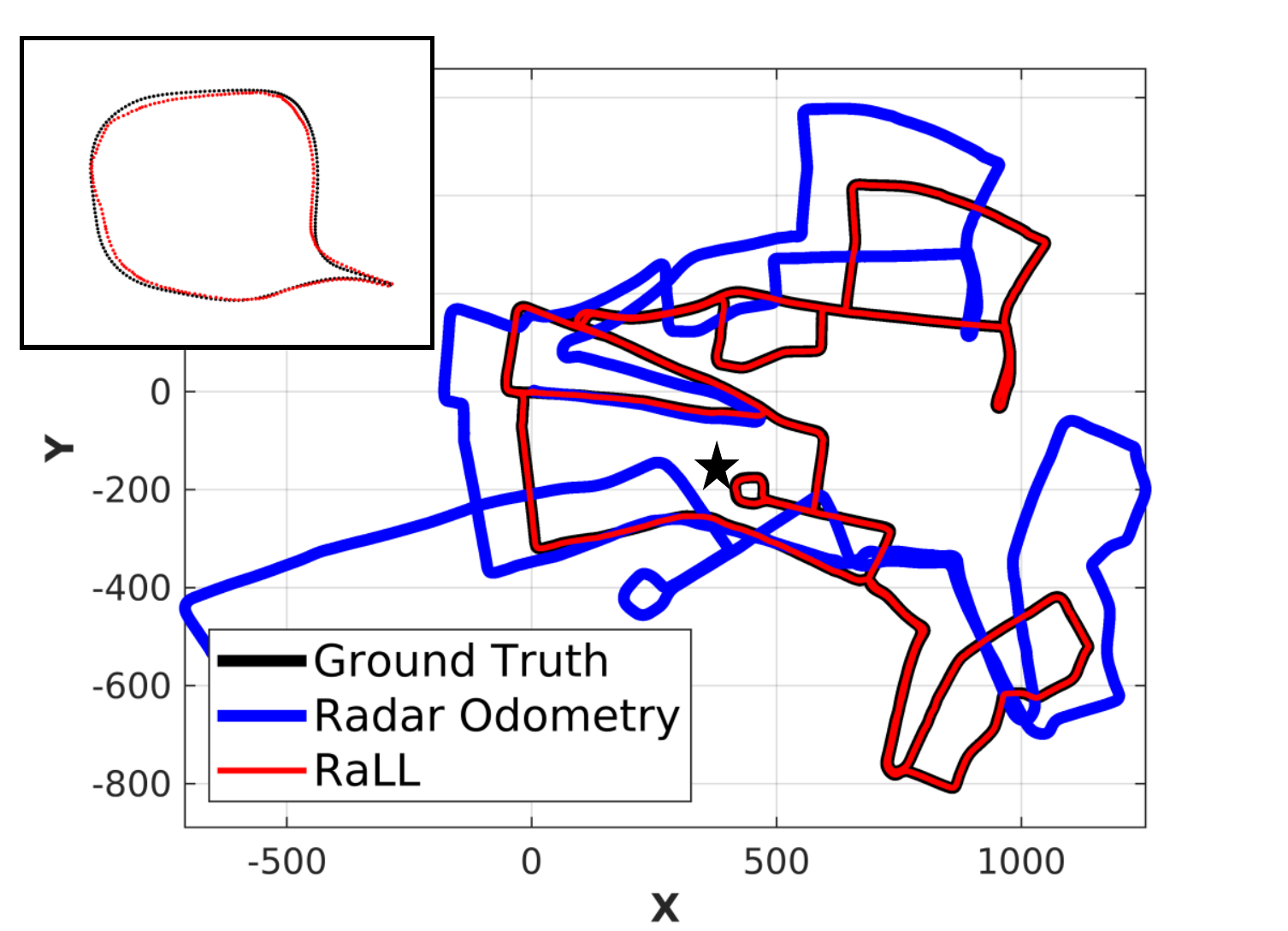}}
	\subfigure[Localization trajectories on Oxford-06]{ \label{Seq-06}	
		\includegraphics[width=5.8cm]{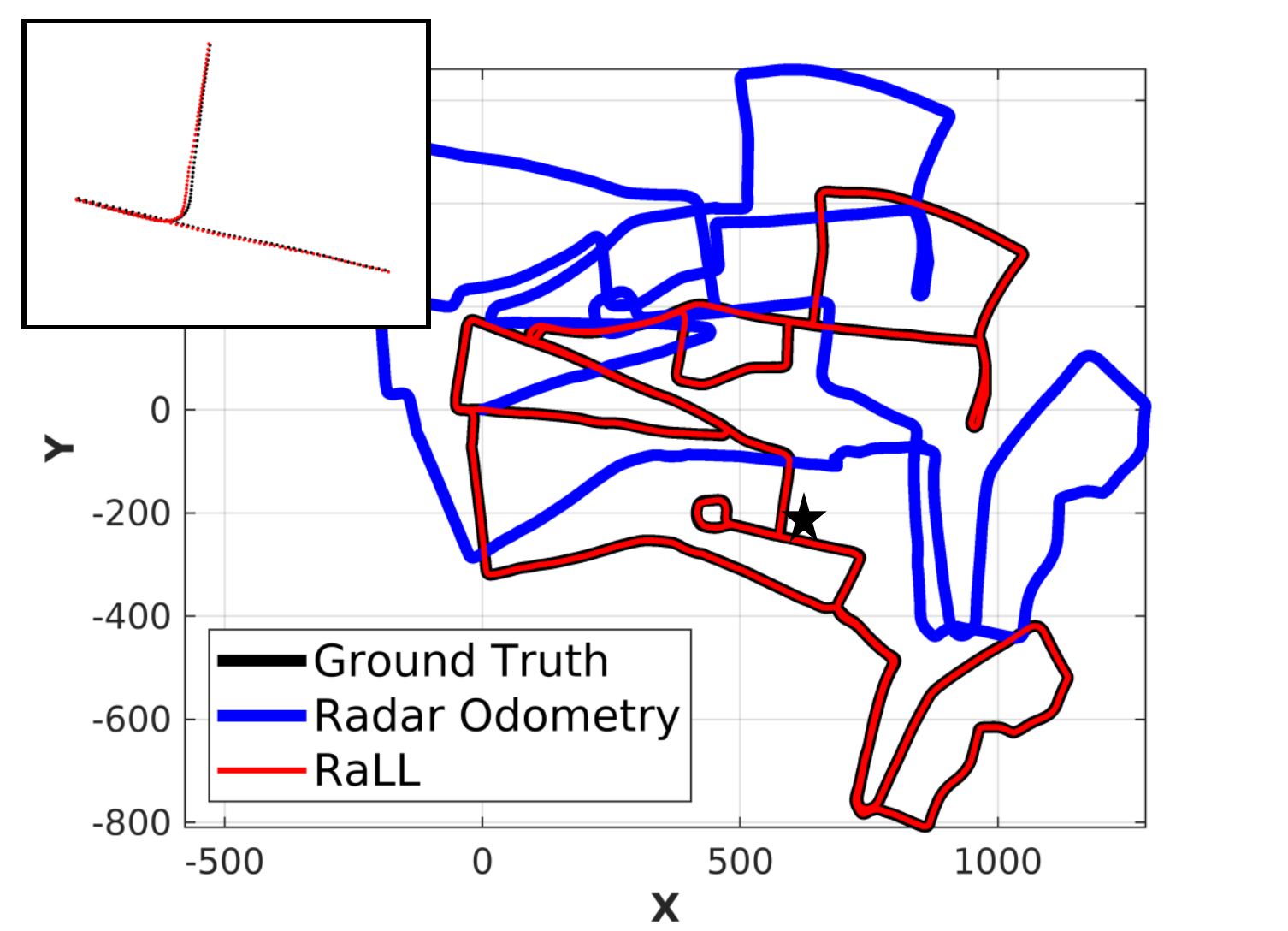}}
	\caption{The localization results on RobotCar dataset. (a) The training and evaluation sections in RobotCar dataset, where the overlapping area is removed. (b)(c) Pose tracking results on Oxford-02 and Oxford-06, including zoomed views at $\star$ locations.}
	\label{trajectories}
\end{figure*}

Table~\ref{RobotCar_Dataset} and Table~\ref{MulRan_Dataset} present the details of the sequences in these two datasets. For RobotCar dataset, we select six sequences at different time, and Oxford-01 is used for mapping and network training. To validate RaLL in this paper, we follow the path split in \cite{tang2020self}, shown in Figure~\ref{train-test}. As for MulRan dataset, sessions at DCC, KAIST and Riverside are used for evaluation. We build the laser maps using DCC-01, KAIST-02 and Riverside-02, and generalize all sequences directly using the learned model in RobotCar. This generalization strategy is performed in all the following experimental sections \ref{sec:ablation} , \ref{sec:tracking} and \ref{sec:offset}.

\subsection{Ablation Study}
\label{sec:ablation}

\begin{figure*}[t]
	\centering
	\subfigure[$\Delta x_m = \Delta y_m = 2m, \Delta \theta_m = 2^\circ$]{	\label{abla_1}
		\includegraphics[width=8.5cm]{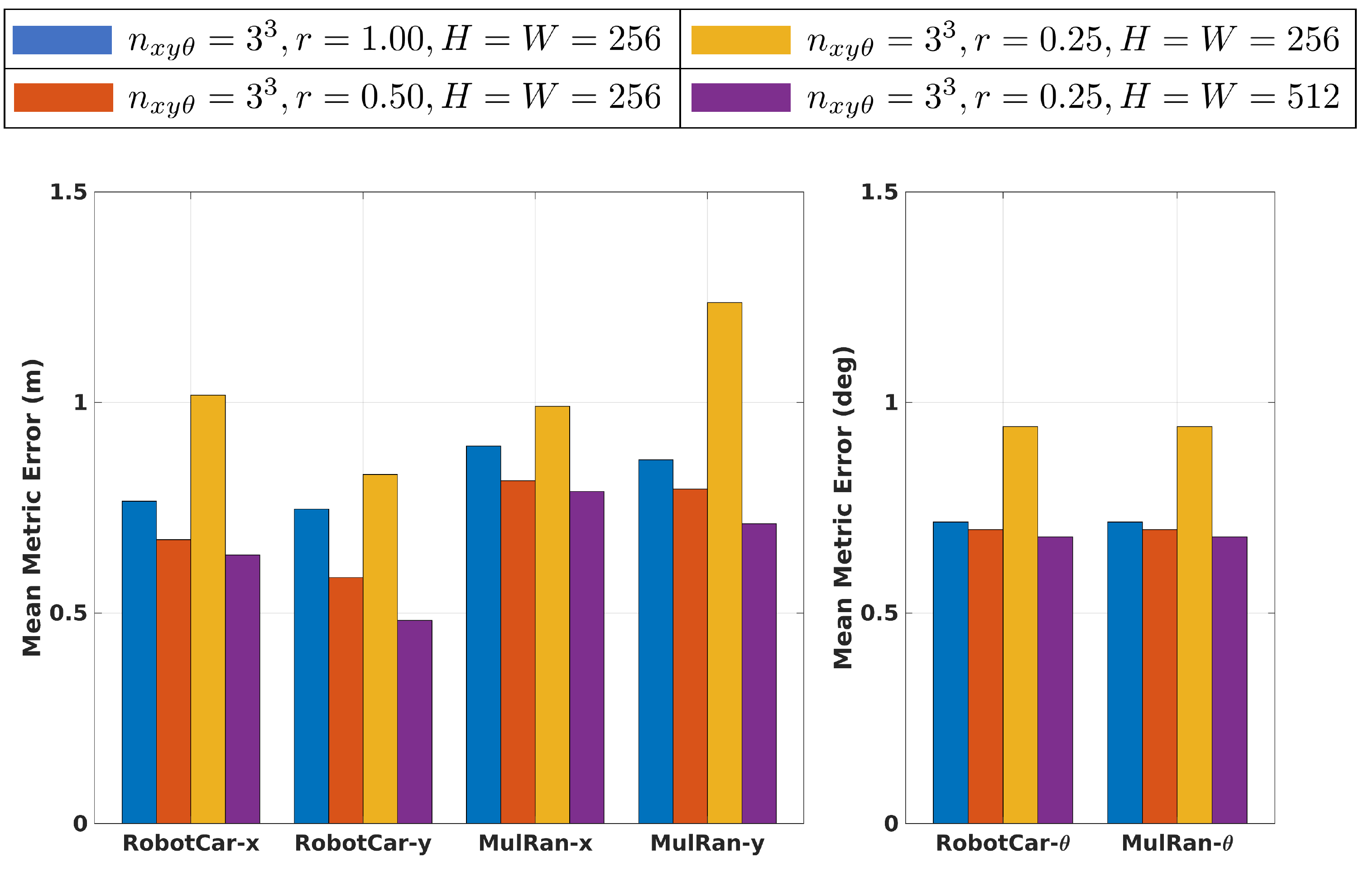}}
	\subfigure[$\Delta x_m = \Delta y_m = 6m, \Delta \theta_m = 6^\circ$]{	\label{abla_2}
		\includegraphics[width=8.5cm]{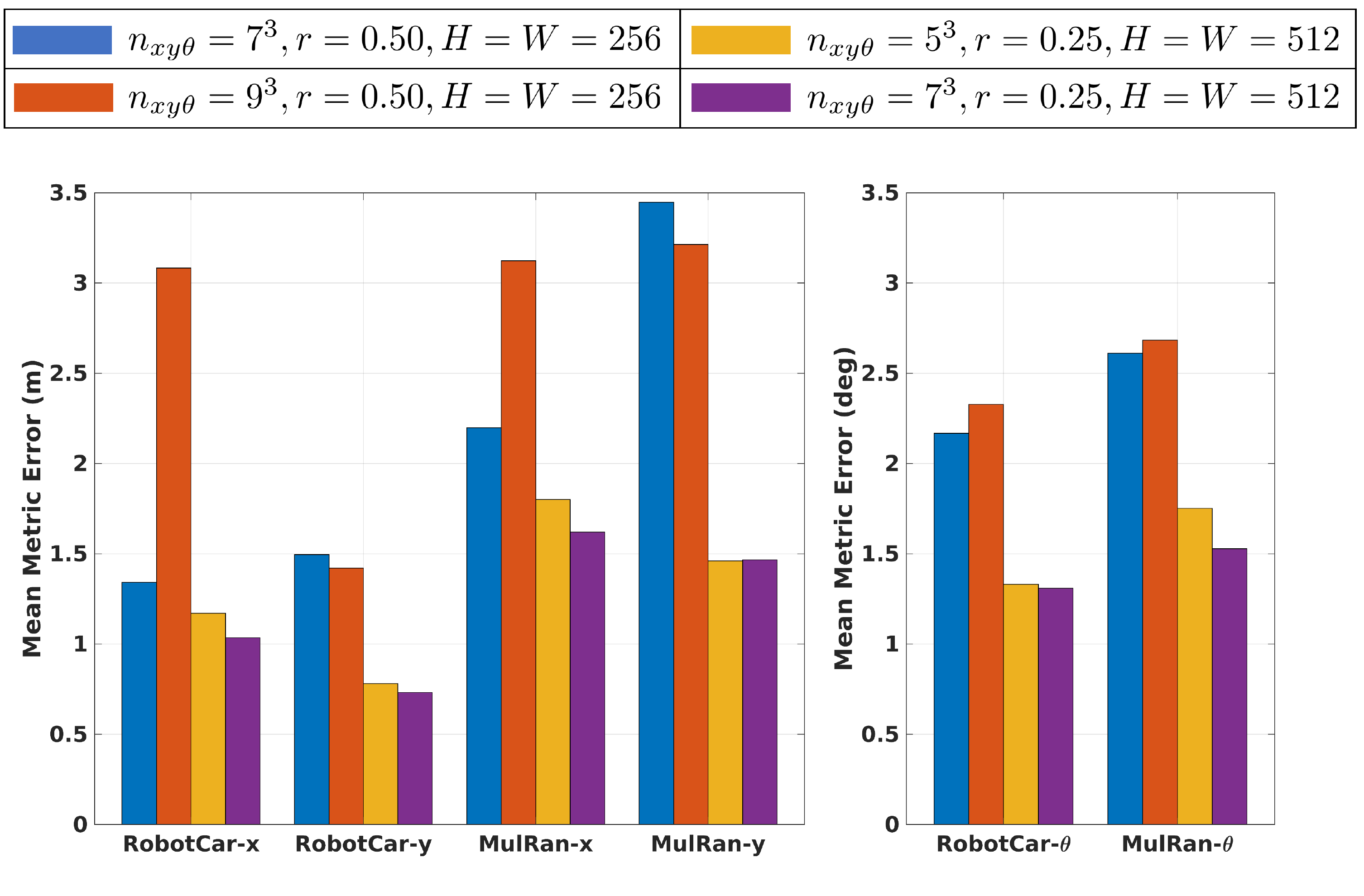}}
	\caption{Ablation study of network configuration parameters.}
	\label{abla}
\end{figure*}

\begin{table*}[t]
	\renewcommand\arraystretch{1.2}
	\begin{center}
		\caption{Ablation study of network architectures}
		\label{ablation_networks}
		\begin{tabular}{p{1.0cm}<{\centering}|p{1.0cm}<{\centering}p{1.0cm}<{\centering}p{1.0cm}<{\centering}|p{1.5cm}<{\centering}p{1.5cm}<{\centering}p{1.5cm}<{\centering}|p{1.5cm}<{\centering}p{1.5cm}<{\centering}p{1.5cm}<{\centering}}
			\hline
			\hline
			\multirow{2}*{No.} & \multicolumn{3}{c|}{Architecture} &\multicolumn{3}{c|}{Mean Metric Error on RobotCar} &\multicolumn{3}{c}{Mean Metric Error on MulRan (Gen.)}\\
			~ &$\mathcal{F}_m$ & $\mathcal{F}_p$ & DKF & $x$(m) & $y$(m) & $\theta\ (^{\circ})$ & $x$(m) & $y$(m) & $\theta\ (^{\circ})$\\
			\hline
			A &  & \cmark &  & 1.06 & 0.80 & 1.44 & 1.65 & 1.42 & 1.55 \\
			B & \cmark &  &  & 1.71 & 1.03 & 2.06 & 2.09 & 2.33 & 2.24 \\
			C & \cmark & \cmark &  & 1.05 & 0.74 & 1.55 & \textbf{1.62} & 1.40 & 1.59 \\
			D & \cmark & \cmark & \cmark & \textbf{1.04} & \textbf{0.69} & \textbf{1.26}  & 1.67  & \textbf{1.37}  & \textbf{1.40} \\
			\hline
			\hline
		\end{tabular}
	\end{center}
\end{table*}

First of all, we employ the sensitivity analysis on the input range images, including the scan size $H,W$ and resolution $r$. Input radar scans and lidar maps share four types of image settings, as shown in Figure~\ref{abla_1}; and then four networks are trained with these different inputs. The parameter $k$ varies when $H=W=256$ or $512$ relatively, thus guaranteeing the invariance of input patch sizes for the patch networks. The ablation study is manual labor consuming for training networks, but we consider it is worthy to explore an appropriate solution for radar localization.

We set the offset in the limit range as follows: $\Delta x_m = \Delta y_m = 2m$, and $\Delta\theta_m = 2^\circ$, and also set the resolution $\delta x= \delta y = 2m, \delta \theta = 2^\circ$ in solution space, thus the cost volume is divided to $n_{xy\theta}=3\times3\times3$. Samples are generated randomly with known offsets for training and evaluation. In RobotCar dataset, we collect more than 1000 samples in the test path of Oxford-02 and Oxford-03; and more than 3000 samples in DCC-02 and KAIST-02 of MulRan dataset. The estimated offset is inferenced by the trained networks $\mathcal{F}_{\mathcal{L}_1 + \mathcal{L}_2}$. We calculate the mean errors of the estimated offsets alongside the three dimensions $x,y,\theta$ in robot coordinate.

The evaluation results are shown in Figure~\ref{abla_1}. It is obvious that the neural networks performs better than the others when $H=W=512$ and $r=0.25m/pixel$. And the network also works well with $H=W=256$ and $r=0.5m/pixel$. In summary, conclusions can be drawn from these comparisons: \emph{the higher range resolution is, or the longer detection range is, the more precise pose estimation will be achieved.} But due to the constrained resources, it is a tradeoff between the image sizes and computing efficiency.

Based on the ablation results above, the best two settings $r=0.5m/pixel$ and $0.25m/pixel$ are selected for further tests. We extend the offset to $\Delta x_m = \Delta y_m = 6m$, and $\Delta\theta_m = 6^\circ$. In this larger space, different resolutions $\delta x,\delta y,\delta \theta$ are tested, resulting in $n_{xy\theta}=7^3$ and $9^3$. Note that the difference of $n_{xy\theta}$ only cause changes of the patch network $\mathcal{F}_p$, in which the number of pose candidates $n_{xy\theta}$ is the number of channels essentially. As shown in Figure~\ref{abla_2}, it is obvious that the neural networks achieve best performance with $n_{xy\theta}=7^3$ and $r=0.25m/pixel$. The comparison between $9^3$ and $7^3$ when $r=0.5m/pixel$ shows that higher resolution in offset space may not bring more precise estimation. We think it is because of the limited neurons in the network, which is hard to learn the distribution of a large number of possibilities.

\begin{table}[!t]
	\renewcommand\arraystretch{1.2}
	\begin{center}
		\caption{Odometric error evaluation on sub-sequences (100m length)}
		\label{odometric_evaluation}
		\begin{tabular}{p{2.8cm}<{\centering}|p{2.75cm}<{\centering}p{2.0cm}<{\centering}}
			\hline
			\hline
			Method & Drift Trans. (\%) & Rot. (deg/m) \\
			\hline
			VO Baseline \cite{churchill2012experience} & 3.9802 & 0.0102 \\
			UnDeepVO \cite{li2018undeepvo} & 4.7683 & 0.0141 \\
			\hline
			Cen RO \cite{2018ICRA_cen} & 3.7168 & 0.0095 \\
			Barnes RO \cite{Barnes2019MaskingByMoving} & 1.1627 & \textbf{0.0030} \\
			Radar SLAM \cite{hong2020radarslam} & 2.1854 & 0.0071 \\
			Fake-Lidar \cite{yin2020radar} & 4.1502 & 0.0119 \\
			RaLL  & \textbf{1.0933} & 0.0037 \\
			\hline
			\hline
		\end{tabular}
	\end{center}
\end{table}

\begin{table}[!t]
	\renewcommand\arraystretch{1.2}
	\begin{center}
		\caption{Relative error evaluation by distance traveled}
		\label{localization_pharao}
		\begin{tabular}{p{2.0cm}<{\centering}|p{2.5cm}<{\centering}p{2.5cm}<{\centering}}
			\hline
			\hline
			\multirow{2}*{Dataset} &\multicolumn{2}{c}{Mean Percentage of Relative Trans. (\%)}  \\
			~ & PhaRaO\cite{park2020pharao} & RaLL \\
			\hline
			DCC-02 & 1.91 & \textbf{0.98} \\ 
			Riverside-02 & 2.36 & \textbf{1.08} \\
			\hline
			\hline
			\multirow{2}*{Dataset} &\multicolumn{2}{c}{Mean Error of Relative Rot. (deg)}  \\
			~ & PhaRaO\cite{park2020pharao} & RaLL \\
			\hline
			DCC-02 & 2.34& \textbf{2.03} \\ 
			Riverside-02 & 4.66& \textbf{1.98} \\
			\hline
			\hline
		\end{tabular}
	\end{center}
\end{table}
Despite the analysis on the configuration paramters, we also conduct the ablation study of the network architectures. To achieve this, we remove or replace some blocks of the original system as follows:
\begin{itemize}
\item $\mathcal{F}_m$: we test the performance with the removal of the masking network $\mathcal{F}_m$, and the radar and lidar feature generation processes are same in this context.
\item $\mathcal{F}_p$: we obtain the similarity scores without the patch division of $\Delta \mathbf{E}$, and difference scores are by applying the networks in Table~\ref{AblationNet}.
\item DKF: we also test the performance with and without the differentiable Kalman Filter, which means the networks are trained with $\mathcal{L}_1 + \mathcal{L}_2$ and $\mathcal{L}_1 + \mathcal{L}_2, \mathcal{L}_3$ respectively. 
\end{itemize}

Table \ref{ablation_networks} shows the ablation study of blocks in system and the experimental results. First, No. C performs better than No. A and B, which demonstrates the effectiveness of the masking network $\mathcal{F}_m$ and patch network $\mathcal{F}_p$. While compared to No. C, No. D is trained with DKF and achieves better performance, validating the superiority of differentiable Kalman Filter. In summary, we validate the effectiveness of our design with the ablation study of system architecture.

\begin{figure*}[t]
	\centering
	\subfigure[Localization trajectories on Riverside-02]{
		\includegraphics[width=5.8cm]{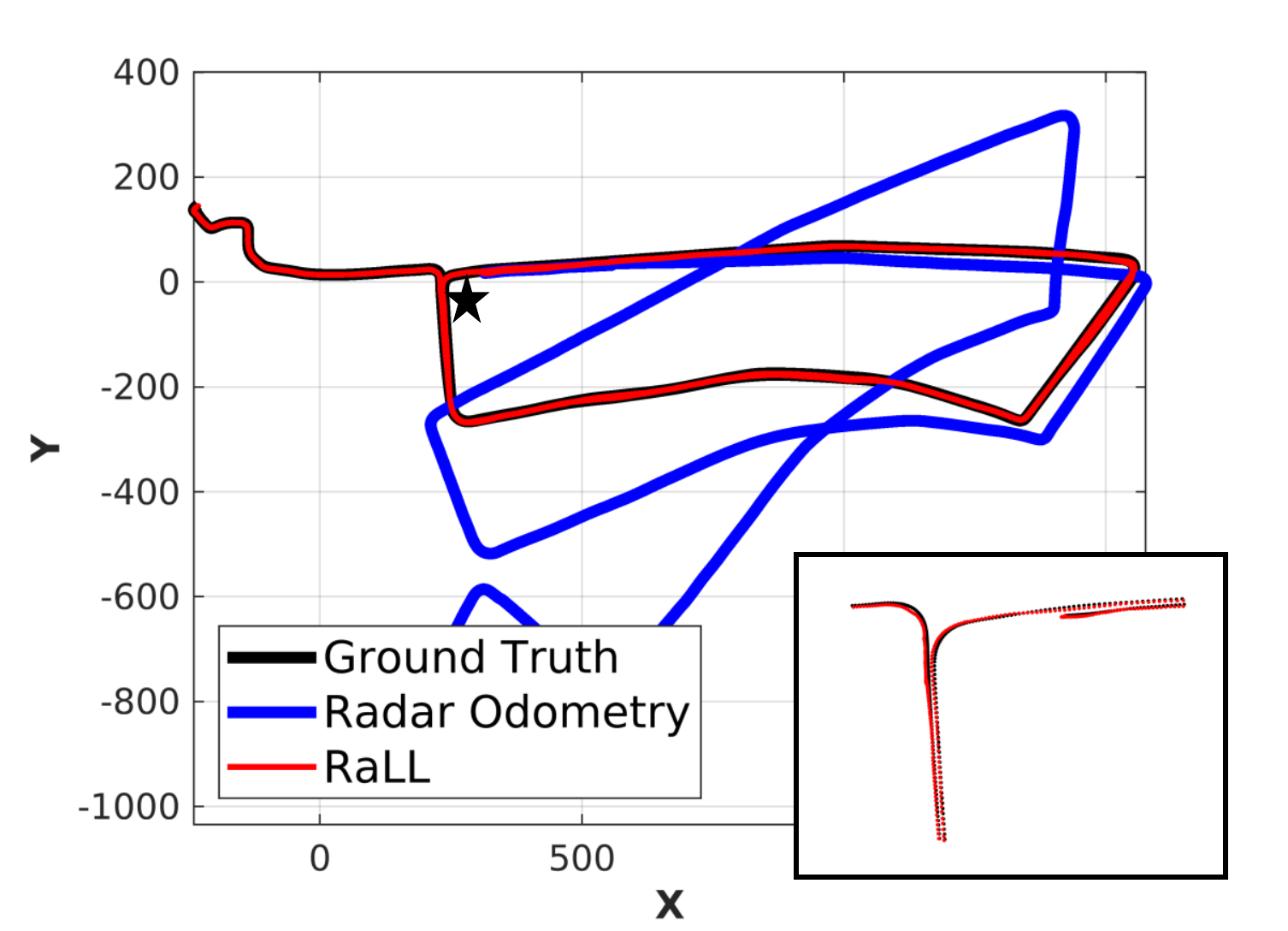}}
	\subfigure[Localization trajectories on DCC-02]{
		\includegraphics[width=5.8cm]{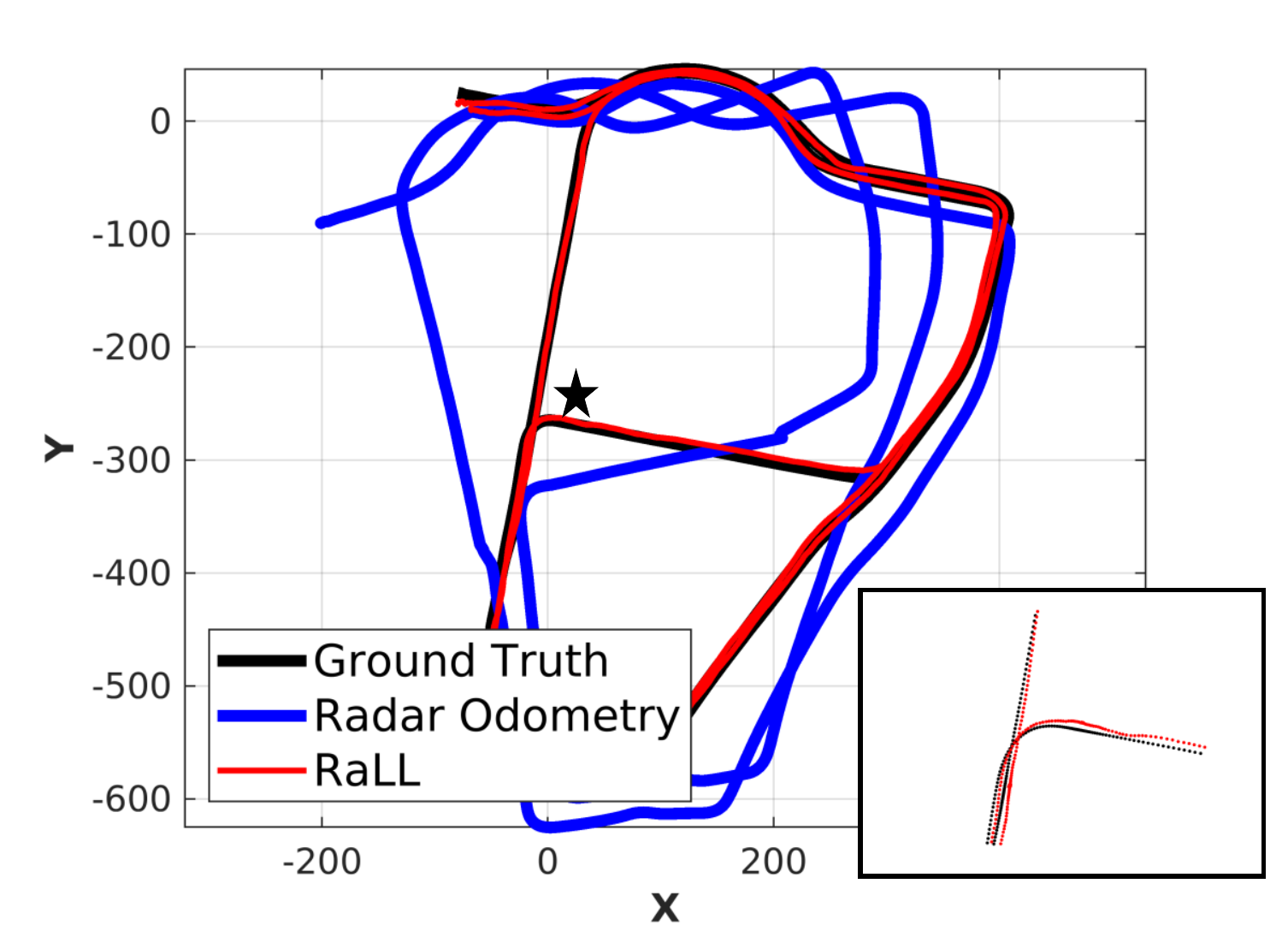}}
	\subfigure[Localization trajectories on KAIST-03]{
		\includegraphics[width=5.8cm]{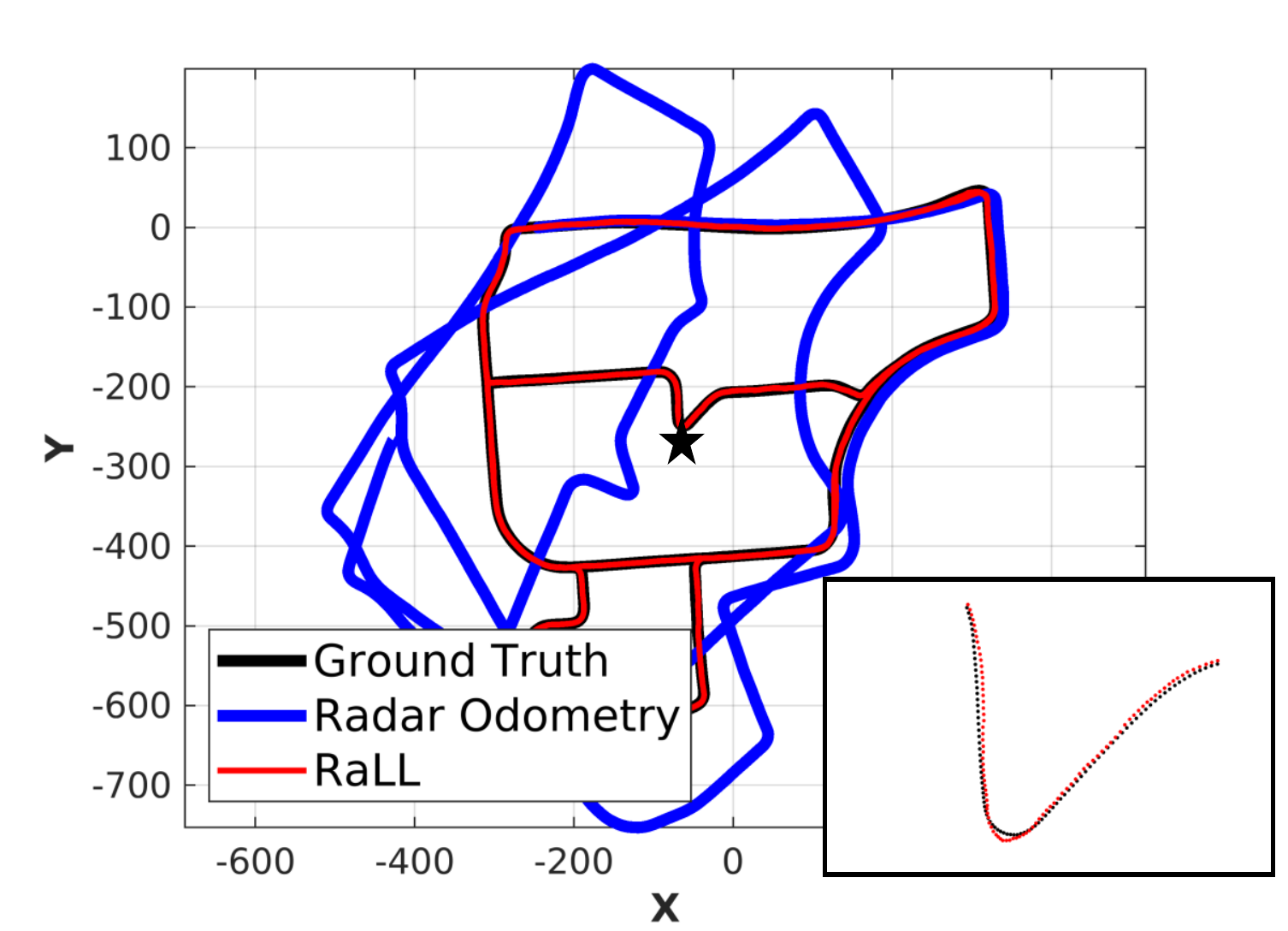}} \\
	
	\subfigure[Relative translation errors on Riverside-02]{
		\includegraphics[width=5.8cm]{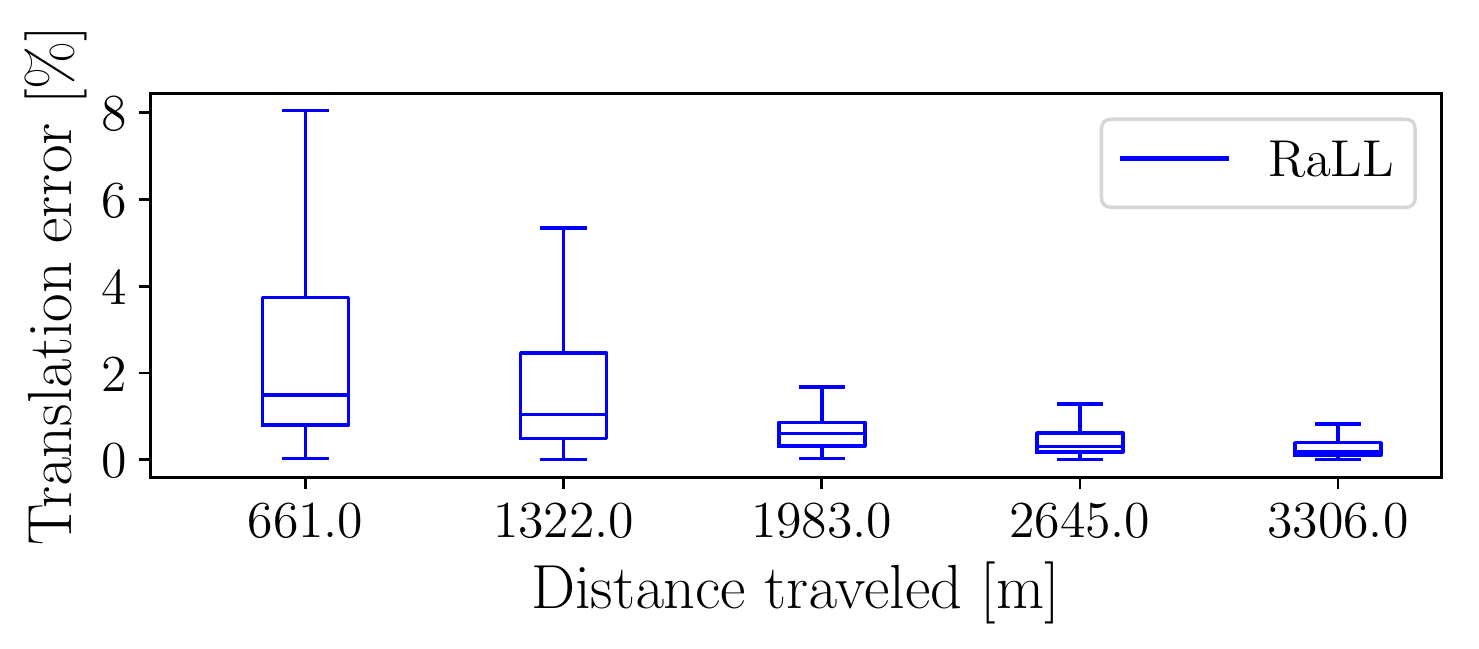}}
	\subfigure[Relative translation errors on DCC-02]{
		\includegraphics[width=5.8cm]{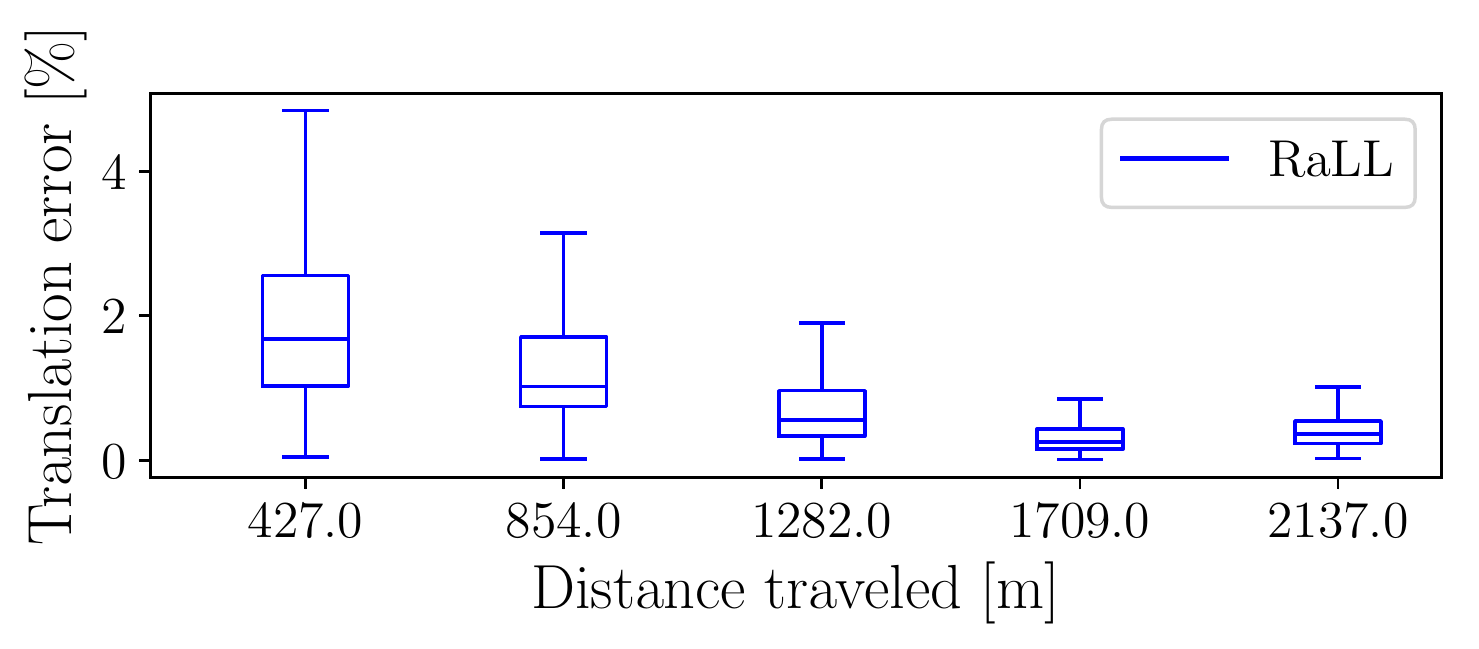}}
	\subfigure[Relative translation errors on KAIST-03]{
		\includegraphics[width=5.8cm]{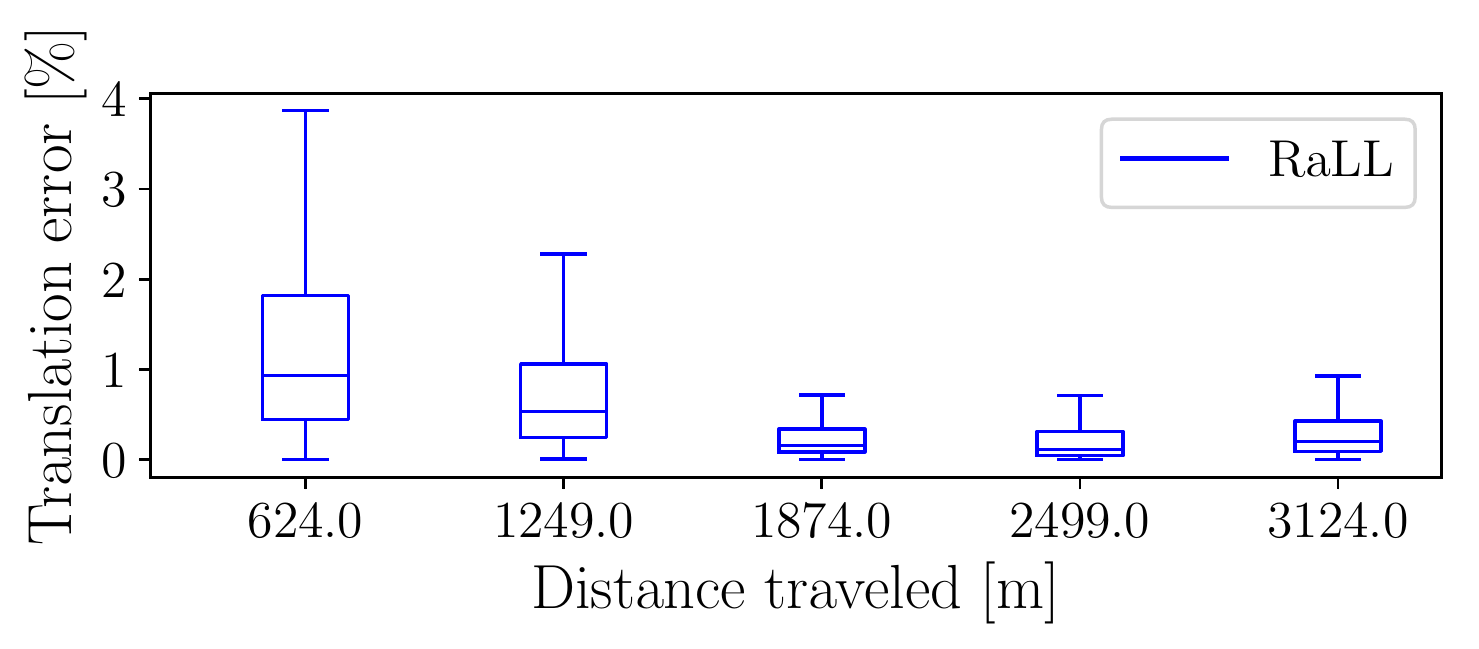}} \\
	
	\subfigure[Relative heading errors on Riverside-02]{
		\includegraphics[width=5.8cm]{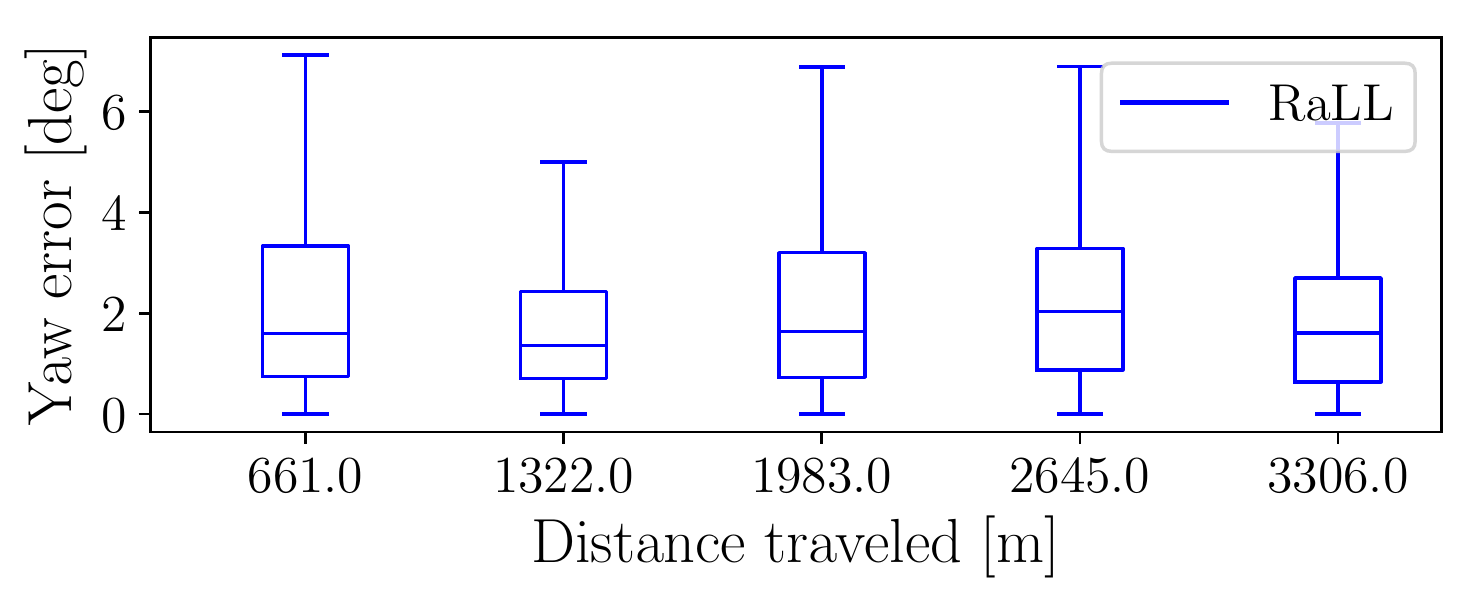}}
	\subfigure[Relative heading errors on DCC-02]{
		\includegraphics[width=5.8cm]{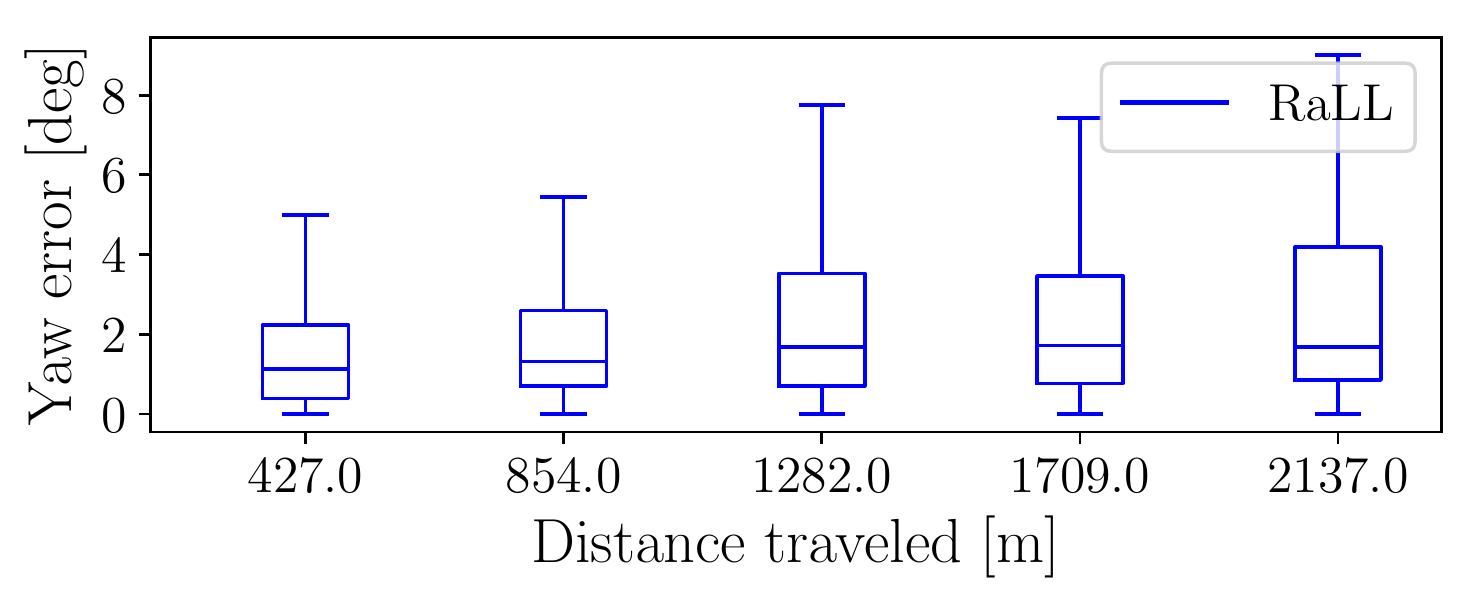}}
	\subfigure[Relative heading errors on KAIST-03]{
		\includegraphics[width=5.8cm]{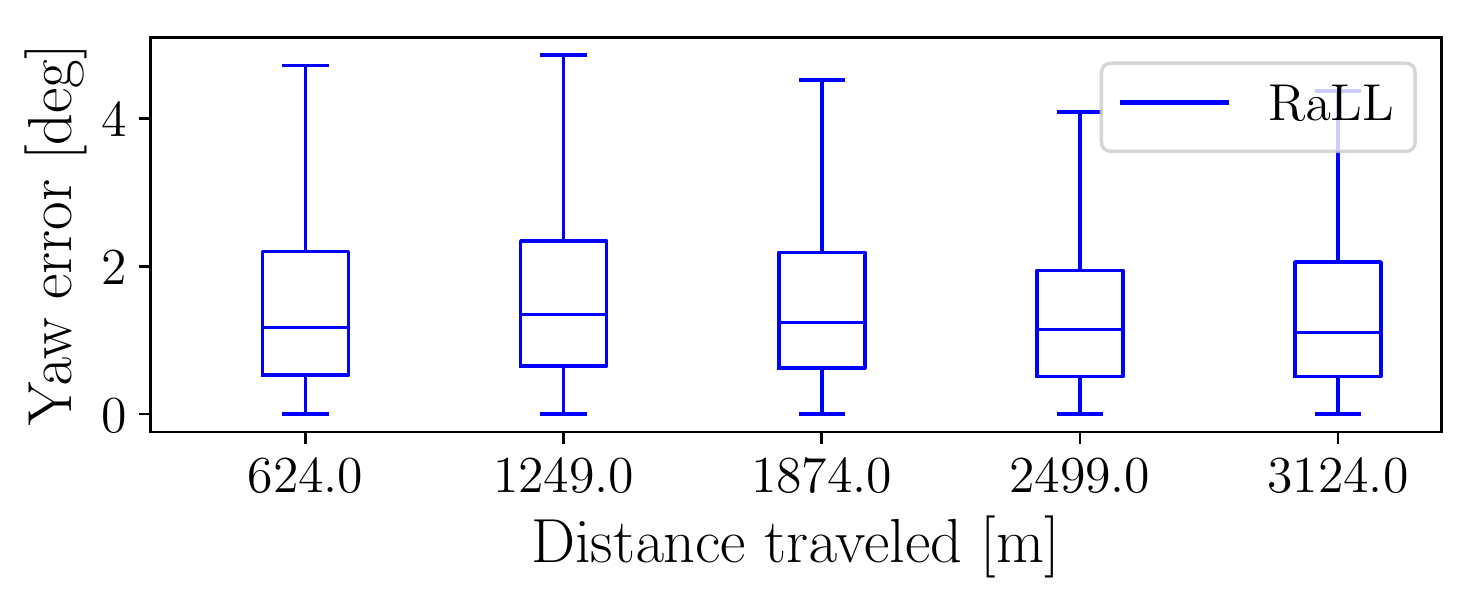}}
	\caption{The localization results on MulRan dataset. (a)(d)(g) The localization trajectories on Riverside-02, percentage of relative translation errors and relative heading errors. (b)(e)(h) reuslts on DCC-02. (c)(f)(i) reuslts on KAIST-03.}
	\label{rel_error_rpg}
\end{figure*}

From these extensive ablation experiments, we also notice that the error in $x$ is generally larger than $y$, which are longitudinal and lateral axes for automative vehicles relatively. We consider this result is interpretable due to the road environments in the two autonomous driving datasets. Most of the roads are straight for vehicles, and are more distinguishable along the lateral direction, especially when there are few features on the road sides. Therefore, it is more difficult to estimate the $\Delta x$ than $\Delta y$ in this paper.

\begin{table*}[t]
	\renewcommand\arraystretch{1.2}
	\begin{center}
		\caption{RMSE of global trajectories}
		\label{localization_rmse}
		\begin{tabular}{p{2.0cm}<{\centering}|p{1.0cm}<{\centering}p{1.0cm}<{\centering}|p{1.1cm}<{\centering}p{1.1cm}<{\centering}|p{1.1cm}<{\centering}p{1.1cm}<{\centering}|p{1.1cm}<{\centering}p{1.1cm}<{\centering}|p{1.1cm}<{\centering}p{1.1cm}<{\centering}}
			\hline
			\hline
			\multirow{2}*{Sequence} & \multicolumn{2}{c|}{Our RO} & \multicolumn{2}{c|}{Fake-Lidar \cite{yin2020radar}} & \multicolumn{2}{c|}{MCL ($\mathcal{F}_{\mathcal{L}_1 + \mathcal{L}_2}$)} &  \multicolumn{2}{c|}{RaLL ($\mathcal{F}_{\mathcal{L}_1 + \mathcal{L}_2}$)} & \multicolumn{2}{c}{RaLL ($\mathcal{F}_{\mathcal{L}_1 + \mathcal{L}_2, \mathcal{L}_3}$)}\\
			~ & Trans.(m) & Rot.($^\circ$) & Trans.(m) & Rot.($^\circ$) & Trans.(m) & Rot.($^\circ$) & Trans.(m) & Rot.($^\circ$) & Trans.(m) & Rot.($^\circ$) \\
			\hline
			Oxford-02& 263.27& 26.93& 8.46& 5.43& 2.96 & 1.89 & 1.42 & 1.53 & \textbf{0.98} & \textbf{1.45}\\
			Oxford-03 & 229.95 & 17.04& 6.93& 2.46& 3.09& 1.92& 1.65 & 1.66 & \textbf{1.14} & \textbf{1.62}\\
			Oxford-04& 131.66& 11.16& 9.12& 4.47& 3.24 & 2.02 & 2.18 & 2.00 & \textbf{1.71} & \textbf{1.93} \\
			Oxford-05& 439.23& 44.14& -& -& 2.93& 1.89& 1.47 & 1.57 & \textbf{1.11} & \textbf{1.48} \\
			Oxford-06& 333.04& 19.45& 14.10& 4.25& 2.96& 1.97& 1.52 & 1.57 & \textbf{1.14} & \textbf{1.52}\\
			RobotCar-all & 303.17& 27.03& 10.06& 4.33& 3.04& 1.94& 1.66& 1.67& \textbf{1.23} & \textbf{1.60}\\
			\hline
			DCC-01& 218.47& 51.41& -& -& 377.08& 106.25& 2.90& 2.01 & \textbf{2.11} & \textbf{1.97}\\
			DCC-02& 96.96& 21.90& -& -& 730.07& 92.45& 5.10& 2.28 & \textbf{4.71} & \textbf{2.01}\\
			DCC-03& 146.28& 33.64& -& -& 7.30& 2.78& 5.88& 2.75 & \textbf{5.14} & \textbf{2.55}\\
			KAIST-01& 376.97& 65.04& -& -& 3.36& 2.06& 1.98 & 1.88 & \textbf{1.45} & \textbf{1.74} \\
			KAIST-02& 216.47& 47.43& -& -& 3.37& 1.99& 1.86 & 1.79 & \textbf{1.30} & \textbf{1.71} \\
			KAIST-03& 242.72& 39.56& -& -& 3.13& 2.00& 1.97 & 1.60 & \textbf{1.27} & \textbf{1.50} \\
			Riverside-01& 367.30& 39.61& -& -& 939.15& 109.86& 497.82 & 48.09 & \textbf{4.12} & \textbf{2.84} \\
			Riverside-02& 407.65& 32.04& -& -& 1245.66& 81.55& 9.52 & 3.19 & \textbf{2.52} & \textbf{1.93} \\
			MulRan-all& 275.61& 41.49& -& -& 610.13& 64.34& 151.35& 14.77& \textbf{3.12} & \textbf{2.02}\\
			\hline
			All sequence& 293.30& 33.11& -& -& 370.42& 39.09& 91.90& 9.09& \textbf{2.13} & \textbf{1.77}\\
			\hline
			\hline
		\end{tabular}
	\end{center}
\end{table*}

\begin{table*}[!htb]
	\renewcommand\arraystretch{1.2}
	\begin{center}
		\caption{Median errors of global trajectories}
		\label{localization_median}
		\begin{tabular}{p{2.0cm}<{\centering}|p{1.0cm}<{\centering}p{1.0cm}<{\centering}|p{1.1cm}<{\centering}p{1.1cm}<{\centering}|p{1.1cm}<{\centering}p{1.1cm}<{\centering}|p{1.1cm}<{\centering}p{1.1cm}<{\centering}|p{1.1cm}<{\centering}p{1.1cm}<{\centering}}
			\hline
			\hline
			\multirow{2}*{Sequence} & \multicolumn{2}{c|}{Our RO} & \multicolumn{2}{c|}{Fake-Lidar \cite{yin2020radar}} & \multicolumn{2}{c|}{MCL ($\mathcal{F}_{\mathcal{L}_1 + \mathcal{L}_2}$)} &  \multicolumn{2}{c|}{RaLL ($\mathcal{F}_{\mathcal{L}_1 + \mathcal{L}_2}$)} & \multicolumn{2}{c}{RaLL ($\mathcal{F}_{\mathcal{L}_1 + \mathcal{L}_2, \mathcal{L}_3}$)}\\
			~ & Trans.(m) & Rot.($^\circ$) & Trans.(m) & Rot.($^\circ$) & Trans.(m) & Rot.($^\circ$) & Trans.(m) & Rot.($^\circ$) & Trans.(m) & Rot.($^\circ$) \\
			\hline
			Oxford-02& 150.61& 19.49& 1.53& 1.10&  2.54& 1.03& 1.07& 0.46& \textbf{0.65} & \textbf{0.40}\\
			Oxford-03 & 207.20& 14.58& 1.38& 1.16&  2.55& 1.06& 1.22& 0.51& \textbf{0.79} & \textbf{0.49}\\
			Oxford-04& 100.48& 8.66& 2.05& 1.32&  2.67& 1.17& 1.62& 0.73& \textbf{1.14} & \textbf{0.68}\\
			Oxford-05& 269.91& 37.77& -& -&  2.50& 1.10& 1.05& 0.59& \textbf{0.71} & \textbf{0.56}\\
			Oxford-06& 283.61& 16.67& 2.04& 1.51&  2.46& 1.09& 1.03& \textbf{0.40}& \textbf{0.88} & 0.43\\
			RobotCar-all& 182.73& 15.77& 1.70& 1.25& 2.54& 1.09& 1.17& 0.53& \textbf{0.80} & \textbf{0.50}\\
			\hline
			DCC-01& 159.41&45.70& -& -& 358.85& 103.19& 2.29& 1.02& \textbf{1.38} & \textbf{0.97}\\
			DCC-02& 92.15& 20.33& -& -& 705.12& 68.83& 4.63& 0.88& \textbf{4.38} & \textbf{0.80}\\
			DCC-03& 92.46& 36.25& -& -& 5.72& 1.60& 5.24& 1.54& \textbf{4.45} & \textbf{1.40}\\
			KAIST-01& 305.38& 47.35& -& -& 2.86& 1.26& 1.70& 1.01& \textbf{1.12} & \textbf{0.93} \\
			KAIST-02& 128.97& 23.22& -& -& 2.79& 1.17& 1.57& 0.86& \textbf{1.05} & \textbf{0.83} \\
			KAIST-03& 174.33& 27.26& -& -& 2.67& 1.24& 1.72& 0.87& \textbf{0.99} & \textbf{0.80} \\
			Riverside-01& 185.04& 31.17& -& -& 745.98& 99.36& 4.76& 1.63& \textbf{2.87} & \textbf{1.32} \\
			Riverside-02& 166.81& 23.52& -& -& 323.36& 29.40& 2.86& 1.33& \textbf{2.15} & \textbf{1.05}\\
			MulRan-all& 141.26& 27.93& -& -& 5.77& 2.45& 2.37& 1.07& \textbf{1.65} & \textbf{0.97}\\
			\hline
			All sequence& 168.78& 18.97& -& -& 3.02& 1.37& 1.52& 0.68& \textbf{1.01} & \textbf{0.65}\\
			\hline
			\hline
		\end{tabular}
	\end{center}
\end{table*}

\subsection{Pose Tracking Evaluation}
\label{sec:tracking}

For the following experiments in Section \ref{sec:tracking}, \ref{sec:dynamic}, \ref{sec:efficiency} and \ref{sec:offset}, we use the best configuration from the ablation study above: $H=W=512, r=0.25m/pixel$ , and the volume is set as $n_{xy\theta}=7^3$ for estimation in $\Delta x_m = \Delta y_m = 6m,\Delta\theta_m = 6^\circ$, and also the proposed differentiable system in this paper.

First, our proposed RaLL is compared to other competitive methods on RobotCar dataset. We follow the experiment settings in \cite{Barnes2019MaskingByMoving} and \cite{hong2020radarslam}, in which the localization precisions were calculated under the odometry evaluation of KITTI benchmarking \cite{Geiger2012CVPR}. Specifically, the errors equal to the mean translation and rotation errors from 100$m$ to 800$m$ with the incremental distance travalled. We use the open source tool \footnote{\href{https://github.com/Huangying-Zhan/kitti-odom-eval}{https://github.com/Huangying-Zhan/kitti-odom-eval}} \cite{zhan2020dfvo} for odometic error calculation on sub-seqeuences.

These competitive methods include the visual odometry based merthods \cite{churchill2012experience, li2018undeepvo}, RO \cite{2018ICRA_cen, Barnes2019MaskingByMoving} and RadarSLAM \cite{hong2020radarslam}. The results are presented in Table~\ref{odometric_evaluation}, in which RaLL achieves the best performance on translation error evaluation, thus validating the effectiveness of the differentiable measurement model and differentiable KF in this paper.

To achieve the extensive comparisons, we also comapre the RaLL with the direct radar odometry PhaRaO \cite{park2020pharao}. We follow the experiment settings in \cite{park2020pharao}, in which the relative errors were computed with the evaluation tool \footnote{\href{https://github.com/uzh-rpg/rpg\_trajectory\_evaluation}{https://github.com/uzh-rpg/rpg\_trajectory\_evaluation}} in \cite{zhang2018tutorial}. In this context, the relative errors are presented in Figure~\ref{rel_error_rpg} with the mean values in Table~\ref{localization_pharao}. Our proposed RaLL performs better than PhaRaO on MulRan dataset though the learned model is trained on RobotCar dataset.

We also calculate the absolute trajectory errors on the whole trajectories. Our proposed radar odometry (RO) is selected as a comparison for evaluation. Furthermore, our previous work \cite{yin2020radar} used GAN to transform radar to fake lidar representations, and MCL was constructed with overlap based mearsurement model. We denote this method as Fake-Lidar. Based on this MCL framework, we also use the differentiable measurement model in this paper for pose tracking, which is denoted as MCL($\mathcal{F}_{\mathcal{L}_1 + \mathcal{L}_2}$). In addition, the particle filter is updated to KF in this paper, which is the proposed RaLL($\mathcal{F}_{\mathcal{L}_1 + \mathcal{L}_2}$) as a comparison. Finally, the complete RaLL with the ($\mathcal{F}_{\mathcal{L}_1 + \mathcal{L}_2, \mathcal{L}_3}$) is performed, which is trained with the supervision of differentiable Kalman Filter.

As shown in Table~\ref{localization_rmse} and \ref{localization_median}, both the Root Mean Squared Errors (RMSE) and median errors of each sequence are calculated with the comparison to the ground truth poses, and we also present the results of the whole dataset. It is obvious that the RaLL($\mathcal{F}_{\mathcal{L}_1 + \mathcal{L}_2, \mathcal{L}_3}$) achieves the best performance over the 90$km$ driving. Several localization trajectries are presented in Figure~\ref{trajectories} and \ref{rel_error_rpg} with the zoomed views.

\begin{figure}[t]
	\centering
	\includegraphics[width=8.5cm]{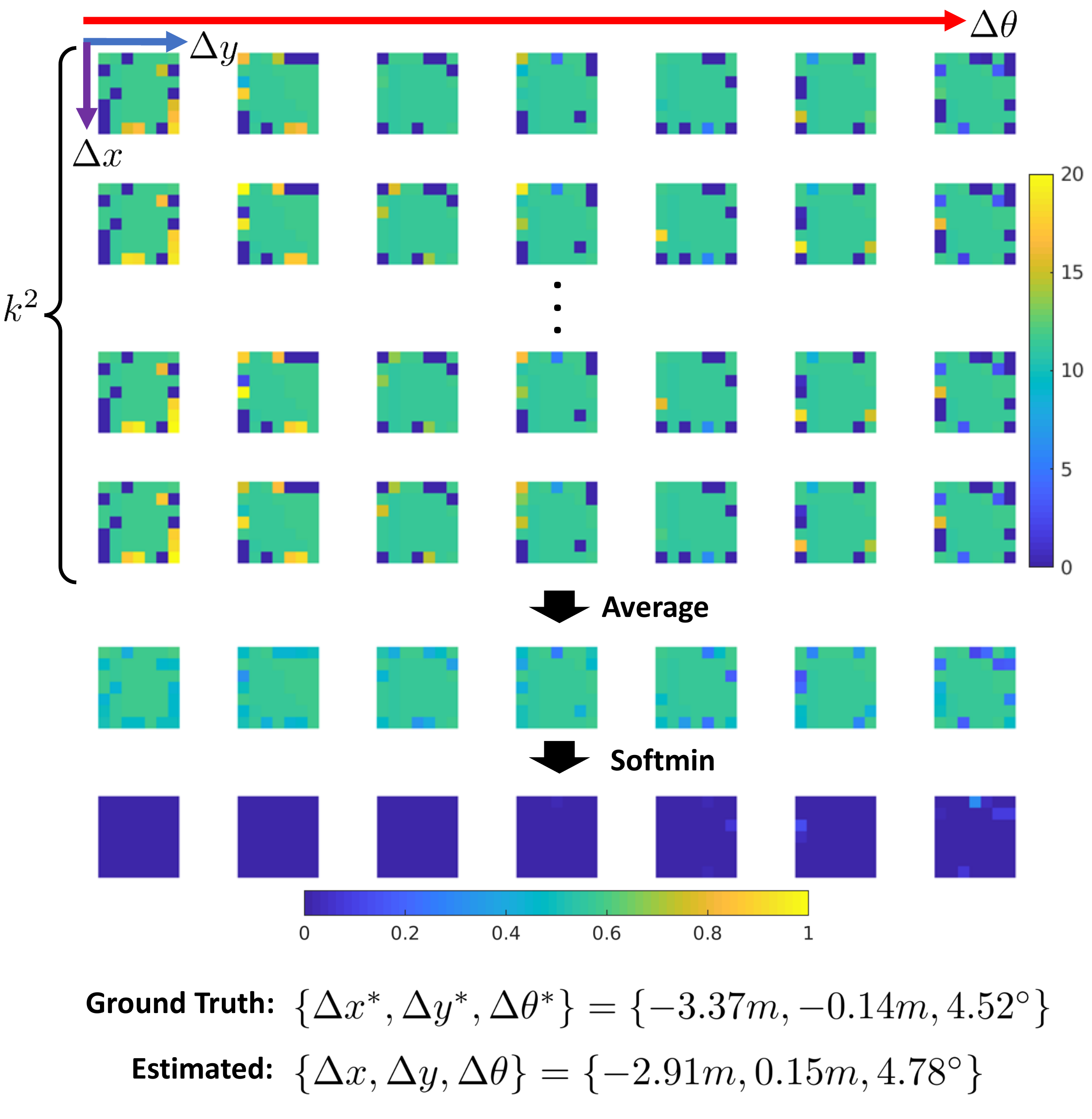}
	\caption{Visualization of the learned probability distribution. Probabilities are expanded along the $\left\lbrace \Delta x, \Delta y, \Delta \theta \right\rbrace$ axes.}
	\label{prob}	
\end{figure}

Figure~\ref{prob} presents a case study on the result in visualization. The resulting tensor size from the patch network $\mathcal{F}_p$ is $(k^2 \times n_{xy\theta} \times 1 \times 1)$, and we formulate it to $\left\lbrace \Delta x, \Delta y, \Delta \theta \right\rbrace$ axes for better visualization. After averaging of $k^2$ tensors and softmin operation, the differences are transfomed to the offset probabilities in the range of $[0,1]$. Finally, the estimated pose offset $\Delta \hat{\mathbf{x}}$ is close to the ground truth $\Delta \mathbf{x}^*$ for pose tracking, and the visualization results demonstrate that our proposed networks have good interpretability.

\subsection{Performance in Dynamic Environments}
\label{sec:dynamic}

Despite the comparisons and evaluations across the whole trajectories, we also investigate the localization performance in dynamic environments. Actually, there are various dynamic objects in both RobotCar and Mulran dataset, including the moving vehicles, bicycles and pedestrians etc. Our proposed RaLL runs successfully in the 90$km$ drives without any localization failure, thus indicating the feasibility of RaLL in dynamic environments.

\begin{figure}[t]
	\centering
	\includegraphics[width=8.5cm]{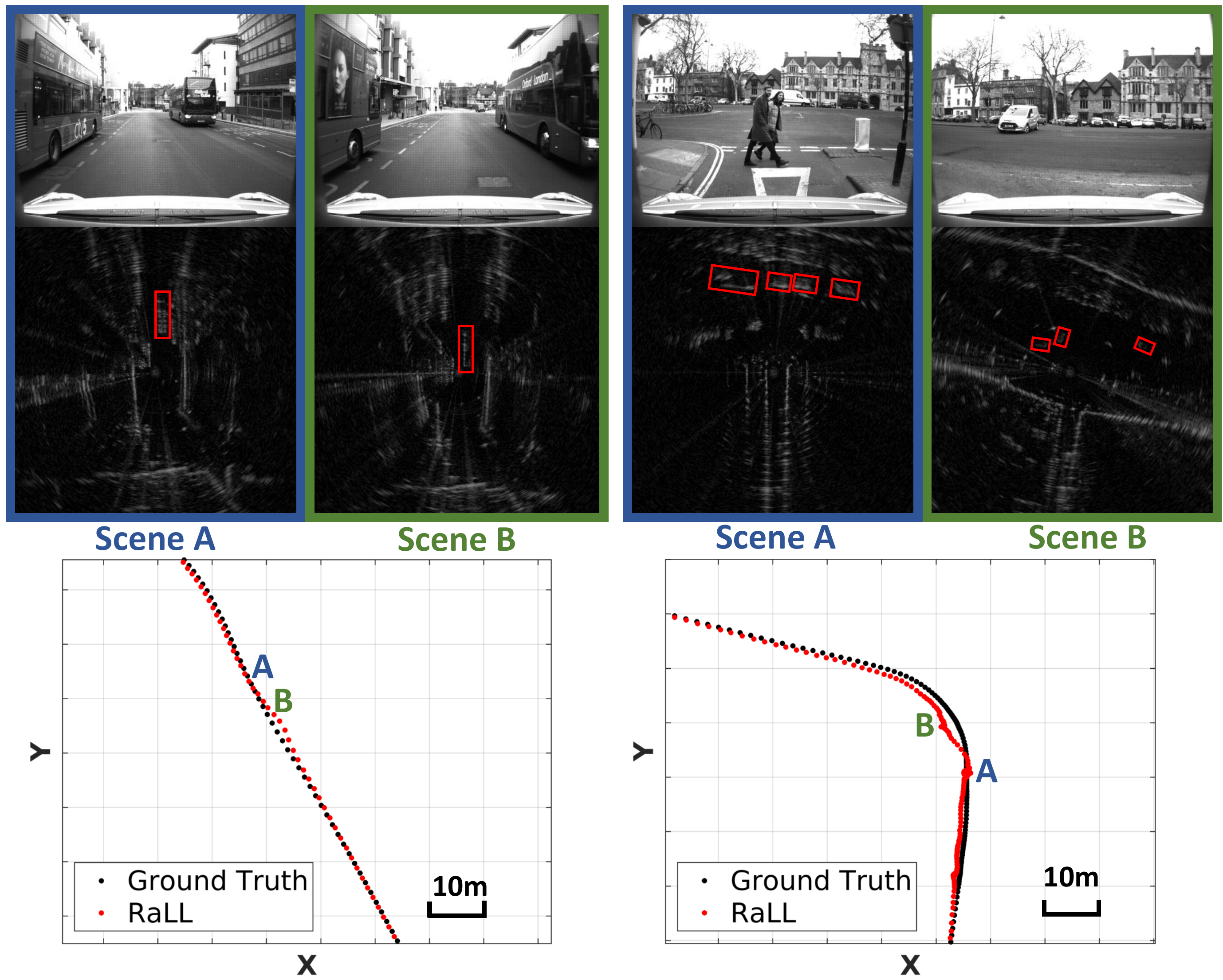}
	\caption{The localization performance in dynamic environments. From left to right: two cases of dynamic environments. From top to bottom: the images from front camera, the radar data from FMCW radar and the trajectories of RaLL and ground truth. Each red bounding box represents the labelled dynamical vehicle.}
	\label{dynamic_env}	
\end{figure}

To better explore the effects of dynamics, we present the peformance in two case studies of Oxford-02, as shown in Figure~\ref{dynamic_env}. Obviously, there are moving objects in the streets and crossings, which have negative effects on our localization precision. However, the RaLL finally corrects the vehicles within limited driving distance. We consider there are two main reasons as follows: first, different from the dynamics in vision \cite{sun2017improving, sun2018motion} and lidar \cite{yin2017efficient, kim2020mulran}, radar sensor suffers less from the object occlusions; second, the proposed measurement model contains the convolution and pooling layers, leading to limited impacts of dynamics in feature embeddings. In summary, our proposed system could be able to handle the dynamics and provide acceptable localization results in dynamical urban scenes.

\begin{table}[t]
	\renewcommand\arraystretch{1.2}
	\begin{center}
		\caption{Timing statistics on each module of RaLL}
		\label{efficiency_rall}
		\begin{tabular}{p{1.4cm}<{\centering}|p{1.2cm}<{\centering}p{1.2cm}<{\centering}p{1.2cm}<{\centering}|p{1.2cm}<{\centering}}
			\hline
			\hline
			Module & Our RO & $\mathcal{F}_{m,r,l,p}$ & KF & Total \\
			\hline
			Time(ms) & 92.78 & 28.90 & 122.19 & 243.87 \\
			\hline
			\hline
		\end{tabular}
	\end{center}
\end{table}

\begin{table}[!t]
	\renewcommand\arraystretch{1.2}
	\begin{center}
		\caption{Efficiencies of recent radar localization methods}
		\label{efficiency_compare}
		\begin{tabular}{p{2.7cm}<{\centering}|p{3.2cm}<{\centering}|p{1.3cm}<{\centering}}
			\hline
			\hline
			Methods & Device & Freq.(Hz) \\
			\hline
			Rapp RO \cite{rapp2017probabilistic} & Intel i7-5930K CPU & 17 \\
			Map Recon. \cite{lu2020see} & Nvidia Jetson TX2 & 1.54 \\
			\hline
			Cen RO \cite{2018ICRA_cen}& / & 3 \\
			Barnes RO \cite{Barnes2019MaskingByMoving} & Nvidia Titan Xp GPU & 13.4 \\
			PhaRaO \cite{park2020pharao} & Intel i7-7700K CPU & 10.73 \\
			Radar SLAM \cite{hong2020radarslam} & Intel i7 CPU & 6 \\
			Our RO & Intel Xeon E7 & 10.78 \\
			RaLL &  Nvidia Titan X GPU & 4.11 \\
			\hline
			\hline
		\end{tabular}
	\end{center}
\end{table}

\subsection{Efficiency Evaluation}
\label{sec:efficiency}

To access the time costs of our proposed radar localization system, we measure the computation efficiency on Oxford-02. More than 8600 radar images are fed into our system and the average time cost of each module is presented in Table~\ref{efficiency_rall}. The total time cost for every inference is 243.87$ms$, and the whole system runs at 4.11 Hz on average. Since the Navtech FMCW radar is operating at 4 Hz, our proposed RaLL achives the real-time performance.

Despite the efficiency analysis of RaLL, we investigate the time costs of other recent research works. Some works are based on learning methods and some are not, leading to the differences on computing devices, so we also list the main computing devices in these papers, as shown in Table~\ref{efficiency_compare}. In addition, some other factors also affect the efficiencies, system implementations and radar resolutions etc, and the final algorithm efficiency may vary in practical applications. 

\textbf{Limitation} For vehicle localization in real world, we consider the transplanting to embedded platform is critically important. The proposed model is deployed on a powerful GPU in this paper, which is an obvious limitation of our proposed RaLL. To overcome this, there are several feasible ways to improve the operating efficiency, such as neural architecture optimization and  multi-gpu parallel programming etc.

\subsection{Localization with the Large Offsets}
\label{sec:offset}

To make more extensive comparisons, we also perform the localization with large initial offset same as \cite{tang2020rsl,tang2020self}, which evaluates the measurement model directly in this paper. For a fair comparison, we expand the solution space to $\Delta x_m = \Delta y_m = 18m, \Delta \theta_m = 18^\circ$, which is close to the experiments in \cite{tang2020rsl,tang2020self}, and is also a very large offset on pixel levels in this paper. More than 200 and 500 samples are generated in the test path of Oxford-02 and KAIST-02.

\begin{figure}[t]
	\centering
	\includegraphics[width=9cm]{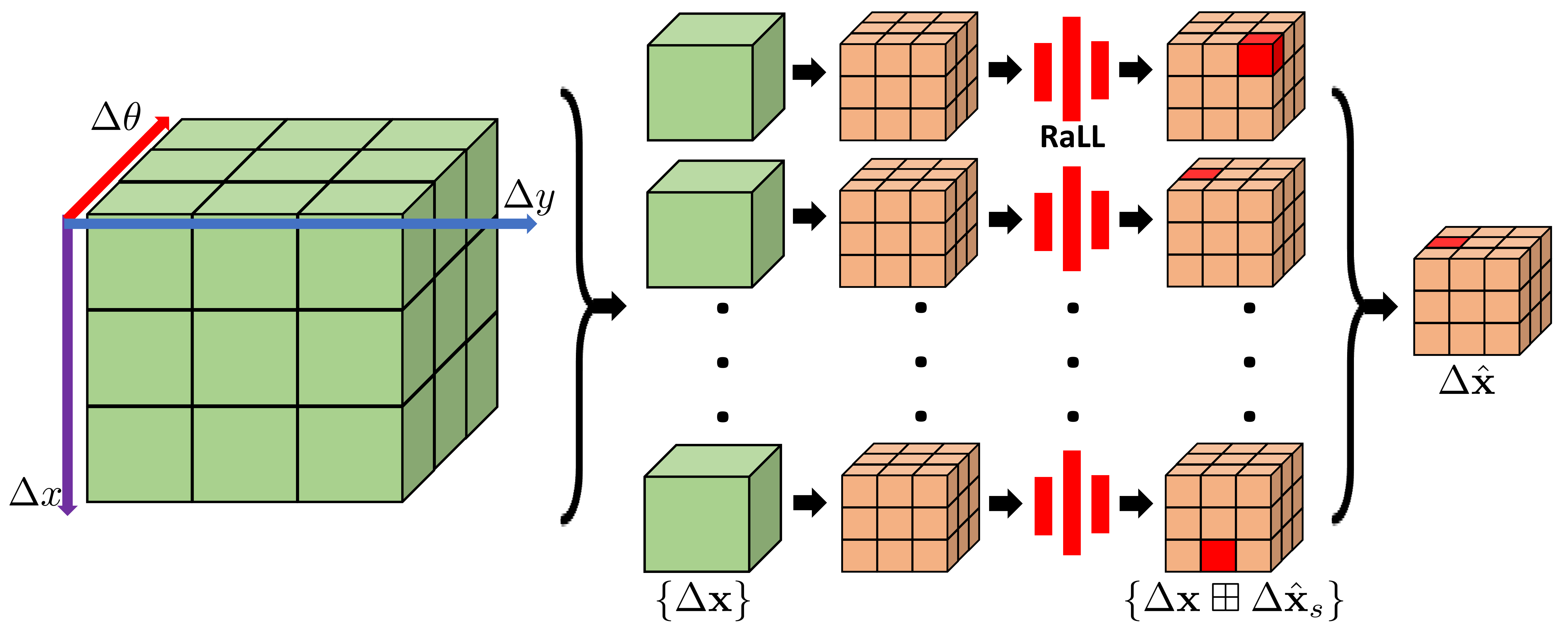}
	\caption{Division of the solution space. Each green sub-space is divided again to obtain the $\Delta \hat{\mathbf{x}}_s$ inside, and the final estimated $\Delta \hat{\mathbf{x}}$ is selected from the from $\left\lbrace \Delta \mathbf{x} \boxplus \Delta \hat{\mathbf{x}}_s\right\rbrace$.  }
	\label{division}	
\end{figure}

However, our proposed network is designed for limited offsets originally, and not applicable to large offsets. To achieve this goal, we divide the entire space to several sub-spaces, illustrated in Figure~\ref{division}. And initial guesses are raised in the center of these sub-spaces as initial offsets $\left\lbrace \Delta \mathbf{x}\right\rbrace $. Then the proposed network is applied at each initial offset, and we can obtain the estimated $\left\lbrace \Delta \hat{\mathbf{x}}_s\right\rbrace $ in sub-spaces. While in the entire space, all the potential solutions are $\left\lbrace \Delta \mathbf{x} \boxplus \Delta \hat{\mathbf{x}}_s\right\rbrace$, where $\boxplus$ is the pose transformation process in Equation~\ref{disturbance}, and the problem is then transformed to find the best solution in this answer set. In Section~\ref{sec:similarity}, a coarse similarity is measured by subtraction of embedding, and here we calculate this similarity to select the best sub-space directly. Finally, in $\left\lbrace \Delta \mathbf{x} \boxplus \Delta \hat{\mathbf{x}}_s\right\rbrace$, we regard the offset with highest similarity as the estimated offset $\Delta \hat{\mathbf{x}}$.

The evaluation results are presented in Table~\ref{compare_large}, and we exchange $x$ and $y$ in \cite{tang2020rsl,tang2020self} because of the different representations in methodologies. Our proposed method performs better overall the performance, compared to another two deep learning based methods. This experiment indicates that our network not only handles long-term pose tracking, but also the localization with large offsets.

\begin{table}[!t]
	\renewcommand\arraystretch{1.2}
	\begin{center}
		\caption{Comparisons of matching with large offsets}
		\label{compare}
		\begin{tabular}{p{1.4cm}<{\centering}p{2.0cm}<{\centering}|p{1.1cm}<{\centering}p{1.1cm}<{\centering}p{1.1cm}<{\centering}}
			\hline
			\hline
			\multirow{2}*{Dataset}& \multirow{2}*{Method} &\multicolumn{3}{c}{Mean Metric Error}  \\
			~ & ~ & $x$ (m)& $y$ (m) & $\theta\ (^{\circ})$  \\
			\hline
			\multirow{3}*{RobotCar}& RSL-Net \cite{tang2020rsl} & 3.41& 2.66& 2.45 \\
			~& PASED \cite{tang2020self} & \textbf{2.57}& 2.21& 2.65 \\
			~& RaLL & 3.51& \textbf{2.05}& \textbf{1.60} \\
			\hline
			\multirow{3}*{MulRan}& RSL-Net \cite{tang2020rsl} & 2.61& 3.31& \textbf{1.40} \\
			~& PASED \cite{tang2020self} & 3.26& 3.57& 2.15 \\
			~& RaLL (Gen.) & \textbf{2.43}& \textbf{2.03}& 1.90 \\
			\hline
			\hline
			\label{compare_large}
		\end{tabular}
	\end{center}
\end{table}

\section{Conclusion}
\label{sec:conclusion}

An end-to-end system RaLL is proposed in this paper, which can localize a rotating radar sensor on a prior lidar map. To achieve this, we first build networks to extract comparable embeddings of different sensor modalities, and then form the pose probabilities via differentiable solver. The Kalman Filter is applied to model the measurement uncertainty in end-to-end manner, and also fuse the measurement and motion for pose tracking. In multi-session multi-scene datasets, we demonstrate the effectiveness of the proposed differentiable localization system, with the comparison to other competitive methods.

On the other hand, there remain some promising directions to overcome the limitations. First, the code transplanting to embedded platform is important for practical applications, and several feasible solutions are expected to improve the operating efficiency, e.g. the multi-core parallel programming. Second, based on the existing end-to-end visual odometry \cite{li2018undeepvo} and lidar odometry \cite{li2019net}, we intend to develop the uncertainty modeling of the radar odometry. Finally, the optimal network model is critical in real applications, which improves both effectiveness and efficiency for deployment, and we consider that the Neural Architecture Search (NAS) \cite{elsken2019neural} can explore the best solution of our network design in the future.

\appendices
\label{sec:appendices}

\section{Network Architecture}
\label{sec:arc}
For the feature embedding network, we follow the general settings in relative works \cite{Barnes2019MaskingByMoving, tang2020self, UnderTheRadarICRA2020}. As for the proposed patch network, we set the input image sizes as $32 \times 32$ for experiments. The abbreviations and network architectures are as follows:
\begin{itemize}
	\item D-Net($C_{in},C_{mid},C_{out}$): A doubled neural network in U-Net
	\item Conv($C_{in},C_{out},k,s,p$): convolution on 2D image with $C_{in}$ input channels, $C_{out}$ output channels, kernel size $k$, stride $s$, and padding $p$
	\item IN: instance normalization
	\item BN: batch normalization
	\item ReLU: rectified linear unit function
	\item Pool($k,s,p$): max pooling with kernel size $k$, stride $s$ and padding $p$
	\item UpSampling($s$): upsample on 2D image with a scale factor $s$
	\item Skip: skip connection between layers
\end{itemize}

The network architectures are shown in Table~\ref{DoubleNet}, Table~\ref{UNet} and Table~\ref{PatchNet}, and Table~\ref{AblationNet} presents the structure for training without the patch network.

\begin{table}[!htb]
	\renewcommand\arraystretch{1.2}
	\begin{center}
		\caption{DNet architecture}
		\begin{tabular}{p{7cm}}
			\hline
			\hline
			Conv($C_{in}$,$C_{mid}$,3,1,1) + IN + ReLU \\
			Conv($C_{mid}$,$C_{out}$,3,1,1) + IN + ReLU \\
			\hline
			\hline
			\label{DoubleNet}
		\end{tabular}
	\end{center}
\end{table}

\begin{table}[!htb]
	\renewcommand\arraystretch{1.2}
	\begin{center}
		\caption{U-Net architecture}
		\begin{tabular}{p{7cm}}
			\hline
			\hline
			D-Net(1,8,8) + Pool(2,2,0) \\
			D-Net(8,16,16) + Pool(2,2,0) \\
			D-Net(16,32,32) + Pool(2,2,0) \\
			D-Net(32,64,64) + Pool(2,2,0) \\
			D-Net(64,64,64) \\
			UpSamping(2) + Skip + D-Net(128,64,32) \\
			UpSamping(2) + Skip + D-Net(64,32,16) \\
			UpSamping(2) + Skip + D-Net(32,16,8) \\
			UpSamping(2) + Skip + D-Net(16,8,8) \\
			\hline
			$\mathcal{F}_m$: Conv(8,1,1,1,0) + Sigmoid \\
			$\mathcal{F}_r$: Conv(8,1,3,1,0) + IN + ReLU \\
			$\mathcal{F}_l$: Conv(8,1,3,1,0) + IN + ReLU \\
			\hline
			\hline
			\label{UNet}
		\end{tabular}
	\end{center}
\end{table}

\begin{table}[!htb]
	\renewcommand\arraystretch{1.2}
	\begin{center}
		\caption{Patch-Net architecture}
		\begin{tabular}{p{7cm}}
			\hline
			\hline
			$\{$ Conv($n_{xy\theta}$,$n_{xy\theta}$,4,2,1) + BN + ReLU $\}$ $\times$ 3 \\
			Conv($n_{xy\theta}$,$n_{xy\theta}$,4,1,0) + BN + ReLU \\
			\hline
			\hline
			\label{PatchNet}
		\end{tabular}
	\end{center}
\end{table}

\begin{table}[!htb]
	\renewcommand\arraystretch{1.2}
	\begin{center}
		\caption{Network without the divided patches}
		\begin{tabular}{p{7cm}}
			\hline
			\hline
			$\{$ Conv($n_{xy\theta}$,$n_{xy\theta}$,4,2,1) + BN + ReLU $\}$ $\times$ 7 \\
			Conv($n_{xy\theta}$,$n_{xy\theta}$,4,1,0) + ReLU \\
			\hline
			\hline
			\label{AblationNet}
		\end{tabular}
	\end{center}
\end{table}

\vspace{-1.5cm}

\bibliographystyle{IEEEtran}
\bibliography{bare_jrnl}

\begin{thebibliography}{10}
\providecommand{\url}[1]{#1}
\csname url@samestyle\endcsname
\providecommand{\newblock}{\relax}
\providecommand{\bibinfo}[2]{#2}
\providecommand{\BIBentrySTDinterwordspacing}{\spaceskip=0pt\relax}
\providecommand{\BIBentryALTinterwordstretchfactor}{4}
\providecommand{\BIBentryALTinterwordspacing}{\spaceskip=\fontdimen2\font plus
\BIBentryALTinterwordstretchfactor\fontdimen3\font minus
  \fontdimen4\font\relax}
\providecommand{\BIBforeignlanguage}[2]{{%
\expandafter\ifx\csname l@#1\endcsname\relax
\typeout{** WARNING: IEEEtran.bst: No hyphenation pattern has been}%
\typeout{** loaded for the language `#1'. Using the pattern for}%
\typeout{** the default language instead.}%
\else
\language=\csname l@#1\endcsname
\fi
#2}}
\providecommand{\BIBdecl}{\relax}
\BIBdecl

\bibitem{mur2015orb}
R.~Mur-Artal, J.~M.~M. Montiel, and J.~D. Tardos, ``Orb-slam: a versatile and
  accurate monocular slam system,'' \emph{IEEE transactions on robotics},
  vol.~31, no.~5, pp. 1147--1163, 2015.

\bibitem{ding2019persistent}
X.~Ding, Y.~Wang, R.~Xiong, D.~Li, L.~Tang, H.~Yin, and L.~Zhao, ``Persistent
  stereo visual localization on cross-modal invariant map,'' \emph{IEEE
  Transactions on Intelligent Transportation Systems}, 2019.

\bibitem{krusi2015lighting}
P.~Kr{\"u}si, B.~B{\"u}cheler, F.~Pomerleau, U.~Schwesinger, R.~Siegwart, and
  P.~Furgale, ``Lighting-invariant adaptive route following using iterative
  closest point matching,'' \emph{Journal of Field Robotics}, vol.~32, no.~4,
  pp. 534--564, 2015.

\bibitem{javanmardi2020pre}
E.~Javanmardi, M.~Javanmardi, Y.~Gu, and S.~Kamijo, ``Pre-estimating
  self-localization error of ndt-based map-matching from map only,'' \emph{IEEE
  Transactions on Intelligent Transportation Systems}, 2020.

\bibitem{lu2019l3}
W.~Lu, Y.~Zhou, G.~Wan, S.~Hou, and S.~Song, ``L3-net: Towards learning based
  lidar localization for autonomous driving,'' in \emph{Proceedings of the IEEE
  Conference on Computer Vision and Pattern Recognition}, 2019, pp. 6389--6398.

\bibitem{barsan2018learning}
I.~A. Barsan, S.~Wang, A.~Pokrovsky, and R.~Urtasun, ``Learning to localize
  using a lidar intensity map.'' in \emph{Conference on Robot Learning (CoRL)},
  2018, pp. 605--616.

\bibitem{carballo2020libre}
A.~Carballo, J.~Lambert, A.~Monrroy, D.~Wong, P.~Narksri, Y.~Kitsukawa,
  E.~Takeuchi, S.~Kato, and K.~Takeda, ``Libre: The multiple 3d lidar
  dataset,'' in \emph{2020 IEEE Intelligent Vehicles Symposium (IV)}.\hskip 1em
  plus 0.5em minus 0.4em\relax IEEE, 2020.

\bibitem{2018ICRA_cen}
S.~H. Cen and P.~Newman, ``Precise ego-motion estimation with millimeter-wave
  radar under diverse and challenging conditions,'' in \emph{2018 IEEE
  International Conference on Robotics and Automation (ICRA)}.\hskip 1em plus
  0.5em minus 0.4em\relax IEEE, 2018, pp. 1--8.

\bibitem{2019ICRA_cen}
------, ``Radar-only ego-motion estimation in difficult settings via graph
  matching,'' in \emph{2019 International Conference on Robotics and Automation
  (ICRA)}.\hskip 1em plus 0.5em minus 0.4em\relax IEEE, 2019, pp. 298--304.

\bibitem{2019ICRA_aldera}
R.~Aldera, D.~De~Martini, M.~Gadd, and P.~Newman, ``Fast radar motion
  estimation with a learnt focus of attention using weak supervision,'' in
  \emph{2019 International Conference on Robotics and Automation (ICRA)}.\hskip
  1em plus 0.5em minus 0.4em\relax IEEE, 2019, pp. 1190--1196.

\bibitem{Barnes2019MaskingByMoving}
D.~Barnes, R.~Weston, and I.~Posner, ``Masking by moving: Learning
  distraction-free radar odometry from pose information,'' in \emph{Conference
  on Robot Learning (CoRL)}, 2020, pp. 303--316.

\bibitem{park2020pharao}
Y.~S. Park, Y.-S. Shin, and A.~Kim, ``Pharao: Direct radar odometry using phase
  correlation,'' in \emph{IEEE International Conference on Robotics and
  Automation (ICRA)}, 2020.

\bibitem{hong2020radarslam}
Z.~Hong, Y.~Petillot, and S.~Wang, ``Radarslam: Radar based large-scale slam in
  all weathers,'' in \emph{2020 IEEE/RSJ International Conference on
  Intelligent Robots and Systems (IROS)}, 2020.

\bibitem{ma2019exploiting}
W.-C. Ma, I.~Tartavull, I.~A. B{\^a}rsan, S.~Wang, M.~Bai, G.~Mattyus,
  N.~Homayounfar, S.~K. Lakshmikanth, A.~Pokrovsky, and R.~Urtasun,
  ``Exploiting sparse semantic hd maps for self-driving vehicle localization,''
  pp. 5304--5311, 2019.

\bibitem{RadarRobotCarDatasetArXiv}
D.~Barnes, M.~Gadd, P.~Murcutt, P.~Newman, and I.~Posner, ``The oxford radar
  robotcar dataset: A radar extension to the oxford robotcar dataset,''
  \emph{arXiv preprint arXiv: 1909.01300}, 2019.

\bibitem{maddern20171}
W.~Maddern, G.~Pascoe, C.~Linegar, and P.~Newman, ``1 year, 1000 km: The oxford
  robotcar dataset,'' \emph{The International Journal of Robotics Research},
  vol.~36, no.~1, pp. 3--15, 2017.

\bibitem{kim2020mulran}
G.~Kim, Y.~S. Park, Y.~Cho, J.~Jeong, and A.~Kim, ``Mulran: Multimodal range
  dataset for urban place recognition,'' in \emph{IEEE International Conference
  on Robotics and Automation (ICRA)}, 2020.

\bibitem{ros2012visual}
G.~Ros, A.~Sappa, D.~Ponsa, and A.~M. Lopez, ``Visual slam for driverless cars:
  A brief survey,'' in \emph{Intelligent Vehicles Symposium (IV) Workshops},
  vol.~2, 2012.

\bibitem{akilan2020multimodality}
T.~Akilan, E.~Johnson, G.~Taluja, J.~Sandhu, and R.~Chadha, ``Multimodality
  weight and score fusion for slam,'' in \emph{2020 IEEE Canadian Conference on
  Electrical and Computer Engineering (CCECE)}.\hskip 1em plus 0.5em minus
  0.4em\relax IEEE, pp. 1--4.

\bibitem{ding2018multi}
X.~Ding, Y.~Wang, H.~Yin, L.~Tang, and R.~Xiong, ``Multi-session map
  construction in outdoor dynamic environment,'' in \emph{2018 IEEE
  International Conference on Real-time Computing and Robotics (RCAR)}.\hskip
  1em plus 0.5em minus 0.4em\relax IEEE, 2018, pp. 384--389.

\bibitem{bresson2017simultaneous}
G.~Bresson, Z.~Alsayed, L.~Yu, and S.~Glaser, ``Simultaneous localization and
  mapping: A survey of current trends in autonomous driving,'' \emph{IEEE
  Transactions on Intelligent Vehicles}, vol.~2, no.~3, pp. 194--220, 2017.

\bibitem{sun2017improving}
Y.~Sun, M.~Liu, and M.~Q.-H. Meng, ``Improving rgb-d slam in dynamic
  environments: A motion removal approach,'' \emph{Robotics and Autonomous
  Systems}, vol.~89, pp. 110--122, 2017.

\bibitem{sun2018motion}
------, ``Motion removal for reliable rgb-d slam in dynamic environments,''
  \emph{Robotics and Autonomous Systems}, vol. 108, pp. 115--128, 2018.

\bibitem{huang2020gmmloc}
H.~Huang, H.~Ye, Y.~Sun, and M.~Liu, ``Gmmloc: Structure consistent visual
  localization with gaussian mixture models,'' \emph{IEEE Robotics and
  Automation Letters}, vol.~5, no.~4, pp. 5043--5050, 2020.

\bibitem{yin20193d}
H.~Yin, Y.~Wang, X.~Ding, L.~Tang, S.~Huang, and R.~Xiong, ``3d lidar-based
  global localization using siamese neural network,'' \emph{IEEE Transactions
  on Intelligent Transportation Systems}, vol.~21, no.~4, pp. 1380--1392, 2019.

\bibitem{park2019radar}
Y.~S. Park, J.~Kim, and A.~Kim, ``Radar localization and mapping for indoor
  disaster environments via multi-modal registration to prior lidar map,'' in
  \emph{2019 IEEE/RSJ International Conference on Intelligent Robots and
  Systems (IROS)}.\hskip 1em plus 0.5em minus 0.4em\relax IEEE, 2019, pp.
  1307--1314.

\bibitem{lu2020see}
C.~X. Lu, S.~Rosa, P.~Zhao, B.~Wang, C.~Chen, J.~A. Stankovic, N.~Trigoni, and
  A.~Markham, ``See through smoke: robust indoor mapping with low-cost mmwave
  radar.'' in \emph{MobiSys}, 2020, pp. 14--27.

\bibitem{vivet2013localization}
D.~Vivet, P.~Checchin, and R.~Chapuis, ``Localization and mapping using only a
  rotating fmcw radar sensor,'' \emph{Sensors}, vol.~13, no.~4, pp. 4527--4552,
  2013.

\bibitem{schuster2016robust}
F.~Schuster, M.~W{\"o}rner, C.~G. Keller, M.~Haueis, and C.~Curio, ``Robust
  localization based on radar signal clustering,'' in \emph{2016 IEEE
  Intelligent Vehicles Symposium (IV)}.\hskip 1em plus 0.5em minus 0.4em\relax
  IEEE, 2016, pp. 839--844.

\bibitem{schuster2016landmark}
F.~Schuster, C.~G. Keller, M.~Rapp, M.~Haueis, and C.~Curio, ``Landmark based
  radar slam using graph optimization,'' in \emph{2016 IEEE 19th International
  Conference on Intelligent Transportation Systems (ITSC)}.\hskip 1em plus
  0.5em minus 0.4em\relax IEEE, 2016, pp. 2559--2564.

\bibitem{rapp2017probabilistic}
M.~Rapp, M.~Barjenbruch, M.~Hahn, J.~Dickmann, and K.~Dietmayer,
  ``Probabilistic ego-motion estimation using multiple automotive radar
  sensors,'' \emph{Robotics and Autonomous Systems}, vol.~89, pp. 136--146,
  2017.

\bibitem{narula2020automotive}
L.~Narula, P.~A. Iannucci, and T.~E. Humphreys, ``Automotive-radar-based 50-cm
  urban positioning,'' in \emph{2020 IEEE/ION Position, Location and Navigation
  Symposium (PLANS)}.\hskip 1em plus 0.5em minus 0.4em\relax IEEE, 2020, pp.
  856--867.

\bibitem{burnett2020we}
K.~Burnett, A.~P. Schoellig, and T.~D. Barfoot, ``Do we need to compensate for
  motion distortion and doppler effects in radar-based navigation?''
  \emph{arXiv preprint arXiv:2011.03512}, 2020.

\bibitem{gadd2020lookaroundyou}
M.~Gadd, D.~De~Martini, and P.~Newman, ``Look around you: Sequence-based radar
  place recognition with learned rotational invariance,'' in \emph{2020
  IEEE/ION Position, Location and Navigation Symposium (PLANS)}.\hskip 1em plus
  0.5em minus 0.4em\relax IEEE, 2020, pp. 270--276.

\bibitem{suaftescu2020kidnapped}
{\c{S}}.~S{\u{a}}ftescu, M.~Gadd, D.~De~Martini, D.~Barnes, and P.~Newman,
  ``Kidnapped radar: Topological radar localisation using
  rotationally-invariant metric learning,'' in \emph{IEEE International
  Conference on Robotics and Automation (ICRA)}, 2020.

\bibitem{UnderTheRadarICRA2020}
D.~Barnes and I.~Posner, ``Under the radar: Learning to predict robust
  keypoints for odometry estimation and metric localisation in radar,'' in
  \emph{IEEE International Conference on Robotics and Automation (ICRA)}, 2020.

\bibitem{tang2020rsl}
T.~Y. Tang, D.~De~Martini, D.~Barnes, and P.~Newman, ``Rsl-net: Localising in
  satellite images from a radar on the ground,'' \emph{IEEE Robotics and
  Automation Letters}, vol.~5, no.~2, pp. 1087--1094, 2020.

\bibitem{tang2020self}
T.~Y. Tang, D.~De~Martini, S.~Wu, and P.~Newman, ``Self-supervised localisation
  between range sensors and overhead imagery,'' in \emph{Robotics: Science and
  Systems (RSS)}, 2020.

\bibitem{yin2020radar}
H.~Yin, Y.~Wang, L.~Tang, and R.~Xiong, ``Radar-on-lidar: metric radar
  localization on prior lidar maps,'' in \emph{2020 IEEE International
  Conference on Real-time Computing and Robotics (RCAR)}.\hskip 1em plus 0.5em
  minus 0.4em\relax IEEE, 2020, best Conference Paper Award.

\bibitem{yin20203d}
H.~Yin, Y.~Wang, L.~Tang, X.~Ding, S.~Huang, and R.~Xiong, ``3d lidar map
  compression for efficient localization on resource constrained vehicles,''
  \emph{IEEE Transactions on Intelligent Transportation Systems}, 2020.

\bibitem{huang2020monocular}
H.~Huang, H.~Ye, Y.~Sun, and M.~Liu, ``Monocular visual odometry using learned
  repeatability and description,'' in \emph{2020 IEEE International Conference
  on Robotics and Automation (ICRA)}.\hskip 1em plus 0.5em minus 0.4em\relax
  IEEE, 2020, pp. 8913--8919.

\bibitem{chen2020survey}
C.~Chen, B.~Wang, C.~X. Lu, N.~Trigoni, and A.~Markham, ``A survey on deep
  learning for localization and mapping: Towards the age of spatial machine
  intelligence,'' \emph{arXiv preprint arXiv:2006.12567}, 2020.

\bibitem{kendall2015posenet}
A.~Kendall, M.~Grimes, and R.~Cipolla, ``Posenet: A convolutional network for
  real-time 6-dof camera relocalization,'' in \emph{Proceedings of the IEEE
  international conference on computer vision}, 2015, pp. 2938--2946.

\bibitem{detone2018superpoint}
D.~DeTone, T.~Malisiewicz, and A.~Rabinovich, ``Superpoint: Self-supervised
  interest point detection and description,'' in \emph{Proceedings of the IEEE
  Conference on Computer Vision and Pattern Recognition Workshops}, 2018, pp.
  224--236.

\bibitem{ronneberger2015u}
O.~Ronneberger, P.~Fischer, and T.~Brox, ``U-net: Convolutional networks for
  biomedical image segmentation,'' in \emph{International Conference on Medical
  image computing and computer-assisted intervention}.\hskip 1em plus 0.5em
  minus 0.4em\relax Springer, 2015, pp. 234--241.

\bibitem{besl1992method}
P.~J. Besl and N.~D. McKay, ``Method for registration of 3-d shapes,'' in
  \emph{Sensor fusion IV: control paradigms and data structures}, vol.
  1611.\hskip 1em plus 0.5em minus 0.4em\relax International Society for Optics
  and Photonics, 1992, pp. 586--606.

\bibitem{paszke2019pytorch}
A.~Paszke, S.~Gross, F.~Massa, A.~Lerer, J.~Bradbury, G.~Chanan, T.~Killeen,
  Z.~Lin, N.~Gimelshein, L.~Antiga \emph{et~al.}, ``Pytorch: An imperative
  style, high-performance deep learning library,'' in \emph{Advances in neural
  information processing systems}, 2019, pp. 8026--8037.

\bibitem{pomerleau2013comparing}
F.~Pomerleau, F.~Colas, R.~Siegwart, and S.~Magnenat, ``Comparing icp variants
  on real-world data sets,'' \emph{Autonomous Robots}, vol.~34, no.~3, pp.
  133--148, 2013.

\bibitem{churchill2012experience}
W.~S. Churchill, ``Experience based navigation: Theory, practice and
  implementation,'' 2012.

\bibitem{li2018undeepvo}
R.~Li, S.~Wang, Z.~Long, and D.~Gu, ``Undeepvo: Monocular visual odometry
  through unsupervised deep learning,'' in \emph{2018 IEEE international
  conference on robotics and automation (ICRA)}.\hskip 1em plus 0.5em minus
  0.4em\relax IEEE, 2018, pp. 7286--7291.

\bibitem{Geiger2012CVPR}
A.~Geiger, P.~Lenz, and R.~Urtasun, ``Are we ready for autonomous driving? the
  kitti vision benchmark suite,'' in \emph{Conference on Computer Vision and
  Pattern Recognition (CVPR)}, 2012.

\bibitem{zhan2020dfvo}
H.~Zhan, C.~S. Weerasekera, J.~Bian, and I.~Reid, ``Visual odometry revisited:
  What should be learnt?'' in \emph{2020 IEEE International Conference on
  Robotics and Automation (ICRA)}.\hskip 1em plus 0.5em minus 0.4em\relax IEEE,
  2020.

\bibitem{zhang2018tutorial}
Z.~Zhang and D.~Scaramuzza, ``A tutorial on quantitative trajectory evaluation
  for visual (-inertial) odometry,'' in \emph{2018 IEEE/RSJ International
  Conference on Intelligent Robots and Systems (IROS)}.\hskip 1em plus 0.5em
  minus 0.4em\relax IEEE, 2018, pp. 7244--7251.

\bibitem{yin2017efficient}
H.~Yin, X.~Ding, L.~Tang, Y.~Wang, and R.~Xiong, ``Efficient 3d lidar based
  loop closing using deep neural network,'' in \emph{2017 IEEE International
  Conference on Robotics and Biomimetics (ROBIO)}.\hskip 1em plus 0.5em minus
  0.4em\relax IEEE, 2017, pp. 481--486.

\bibitem{li2019net}
Q.~Li, S.~Chen, C.~Wang, X.~Li, C.~Wen, M.~Cheng, and J.~Li, ``Lo-net: Deep
  real-time lidar odometry,'' in \emph{Proceedings of the IEEE Conference on
  Computer Vision and Pattern Recognition}, 2019, pp. 8473--8482.

\bibitem{elsken2019neural}
T.~Elsken, J.~H. Metzen, F.~Hutter \emph{et~al.}, ``Neural architecture search:
  A survey.'' \emph{J. Mach. Learn. Res.}, vol.~20, no.~55, pp. 1--21, 2019.

\end{thebibliography}

\end{document}